\documentclass[twoside]{article}

\usepackage{PRIMEarxiv_merged}

\usepackage[T1]{fontenc}

\usepackage{hyperref}
\usepackage{url}
\usepackage{booktabs}
\usepackage{amsfonts}
\usepackage{nicefrac}
\usepackage{microtype}
\usepackage{graphicx}
\usepackage{multirow}
\usepackage{caption}
\usepackage{subcaption}
\usepackage{placeins}
\usepackage{bm}
\usepackage{makecell}
\usepackage{pifont}
\usepackage{xcolor}
\usepackage{tikz}
\usepackage{cleveref}

\newcommand{\Real}{\mathbb{R}}

\newcommand{\G}{\mathcal{G}}
\newcommand{\V}{\mathcal{V}}
\newcommand{\E}{\mathcal{E}}

\newcommand{\subheader}[1]{\noindent\textbf{#1}}
\newcommand{\capheader}[1]{\textit{\textbf{#1}}}

\newcommand{\dygformer}{\emph{DyGFormer}}
\newcommand{\tgat}{\emph{TGAT}}
\newcommand{\jodie}{\emph{JODIE}}
\newcommand{\dyrep}{\emph{DyRep}}
\newcommand{\cawn}{\emph{CAWN}}
\newcommand{\mixer}{\emph{GraphMixer}}
\newcommand{\tgn}{\emph{TGN}}
\newcommand{\tcl}{\emph{TCL}}

\newcommand{\sig}{\textsuperscript{\dag}}

\newlength{\groupgap}
\newcolumntype{G}{@{}p{\groupgap}@{}}

\newcommand{\cmark}{\textcolor{green!60!black}{\ding{51}}} 
\newcommand{\xmark}{\textcolor{red!70!black}{\ding{55}}}   
\newcommand{\tmark}{\textcolor{yellow!40!orange}{$\boldsymbol{\sim}$}} 

\newcommand*\circled[1]{\tikz[baseline=(char.base)]{
		\node[shape=circle,draw,inner sep=1pt] (char) {#1};}}

\title{What Do Temporal Graph Learning Models Learn? 
}

\author{
	Abigail J. Hayes\thanks{Equal contribution. Author order among the co-first authors may be adjusted for individual use.} \\
	University of Mannheim \\
	\texttt{abigail.hayes@uni-mannheim.de} 
	\And
	Tobias Schumacher$^*$ \\
	University of Mannheim, \\
	RWTH Aachen University \\
	\texttt{tobias.schumacher@uni-mannheim.de} 
	\AND
	Markus Strohmaier \\
	University of Mannheim,\\
	GESIS - Leibniz Institute for the Social Sciences, and\\
	Complexity Science Hub Vienna \\
	\texttt{markus.strohmaier@uni-mannheim.de}
}

\begin{document}
	
	\maketitle

	\begin{abstract}
		Learning on temporal graphs has become a central topic in graph representation learning, with numerous benchmarks indicating the strong performance of state-of-the-art models. However, recent work has raised concerns about the reliability of benchmark results, noting issues with commonly used evaluation protocols and the surprising competitiveness of simple heuristics. 
		This contrast raises the question of which characteristics of the underlying graphs temporal graph learning models actually use to form their predictions.
		We address this by systematically evaluating eight models on their ability to capture eight fundamental characteristics related to the link structure of temporal graphs. These include structural characteristics such as density, temporal patterns such as recency, and edge formation mechanisms such as homophily. Using both synthetic and real-world datasets, we analyze how well models learn these characteristics. Our findings reveal a mixed picture: models capture some characteristics well but fail to reproduce others. With this, we expose important limitations. Overall, we believe that our results provide practical insights for the application of temporal graph learning models and motivate more interpretability-driven evaluations in graph learning research.
	\end{abstract}

	\section{Introduction}
	
	Learning on temporal 
	graphs has become an increasingly popular research topic, exemplified by the emergence of a multitude of benchmarks (e.g., \citet{huang_temporal_2023, huang_benchtemp_2024, gastinger_tgb_2024}), on which state-of-the-art graph learning models often appear to achieve very strong results.
	At the same time, the benchmark performances on link prediction tasks have faced increased scrutiny: from flaws in test sets and evaluation metrics leading to unrealistic results \citep{poursafaei_better_2022, li_evaluating_2023, cornell_are_2025}, to heuristics, such as  predicting edges involving recently active and globally popular nodes \citep{cornell_power_2025}, performing on a par with state-of-the-art models. Further, even if specific models perform well on given benchmarks, it is not clear which factors contribute to this or, more specifically, which graph characteristics models pick up on to form their predictions.
	
	In light of these issues with the evaluation of link prediction, we aim to step back and evaluate the ability of popular graph learning models to learn simple, interpretable characteristics of temporal graphs.
	Specifically, we evaluate the ability of dynamic graph learning models to learn eight different characteristics: the general graph characteristics of \emph{temporal granularity}, \emph{edge direction} and \emph{density}, temporal patterns with \emph{edge persistence}, \emph{periodicity} and \emph{recency}, and edge formation mechanisms with \emph{homophily} and \emph{preferential attachment}.
	Success at learning these characteristics is tested for eight state-of-the-art temporal graph learning models, using a range of empirical and synthetic test datasets, totaling over 6,500 models trained for our experiments.
	Our results, summarized in Table \ref{tab:results_summary}, illustrate both strengths and some striking limitations of popular state-of-the-art models, and provide insights for the practical application of graph learning models.
	We believe that our work can increase understanding and inspire more interpretability-focused evaluations of temporal graph learning models.

	\section{Background}
	Before describing our experiments, we introduce key concepts of our work and summarize related research.
	
	\subsection{Preliminaries}
	We first establish the temporal graph setting and the link prediction-based learning framework used throughout the rest of this paper.
	
	\subheader{Temporal Graphs.} 
	In literature, there are two main approaches to modeling temporal networks, using either continuous or discrete time \citep{zheng_survey_2025, huang_utg_2025}.
	
	In the continuous-time setting, the temporal graph can be considered as a stream of edges with fine-grained timestamps. A graph $\G$ can be modeled as a tuple $\G = (\V, \E)$, where $\V = \{1,\dots,N\}$ denotes the set of all nodes, $\E = \left\{ (u_i, v_i, t_i)_{i\in\{1,2\dots\}}\right\}$ the set of edge events, $u_i, v_i\in\mathcal{V}$ the source and destination nodes respectively, and $t_i$, with $t_i\leq t_{i+1}\, \forall i$, the timestamp of an edge event.
	
	For the discrete-time setting, a temporal graph is instead considered as a series of static graph snapshots with a single time step representing a longer time period. A graph $\G$ is typically modeled as a sequence of graph snapshots $\G = (\G_t)_{t\in\{1,2\dots T\}}$, where each snapshot corresponds to a tuple $\G_t = (\V_t, \E_t)$ with $\V_t$ denoting the nodes, $\E_t$ the edges at time $t$, and $T$ the total number of snapshots.
	
	Datasets can be transformed between the two graph settings, but transformation from continuous to discrete-time typically causes some information loss when binning highly granular timestamps to discrete time steps \citep{huang_utg_2025}.
	For the scope of this work, we focus on models designed for continuous-time dynamic graphs, as these are now more commonly used within the research community. However, we will often consider datasets with discretized time steps due to our specific experimental designs.
	Within the scope of this work, unless otherwise specified, we generally ignore the use of edge and node features to simplify notation.

	\subheader{Representation Learning for Temporal Link Prediction.}
	Most state-of-the-art graph learning models for dynamic graphs are based on various kinds of neural networks, from recurrent neural networks \citep{kumar_predicting_2019}, to graph neural networks \citep{rossi_temporal_2020} and graph transformers \citep{yu_better_2024}.
	During training, continuous-time models receive the sequence of edges $\E$ up to a time $t$ as input and learn a time-aware representation $\bm{z}_v^t\in\Real^D$ for each node $v\in\V$. 
	For dynamic link prediction, the input edges are directly used as positive examples, and optimized such that the existence of an edge $(u,v)$ at test time $t'\geq t$ can be predicted from the corresponding pair of representations $\bm{z}^t_u, \bm{z}^t_v$.
	There is some technical complexity in this optimization, as most models also require negative, i.e., non-existent edges, in training.
	Using all negative edges is generally infeasible since the size of the training data grows quadratically with the number of nodes. Further, empirical social networks are typically very sparse, such that models would be prone to predicting all edges as negative.
	Instead, a set of negative training edges needs to be sampled \citep{yang_evaluating_2015}. 
	This is further complicated in the temporal context by also needing to consider issues such as the timestamps of negative samples, whether or not to specifically sample edges which have been present previously and whether to include nodes which will only become active at a future timestamp.
	Similar issues also carry over to the model evaluation, where negative samples are incorporated in any test set. In the literature, it is a common approach to sample one negative edge for each positive edge at the same timestamp, but there are more complex strategies, such as explicitly sampling from previously positive edges \citep{yu_better_2024}.

	\subsection{Related Work}
	
	We briefly review the work most closely related to the motivation and evaluation of our approach.
	
	\subheader{Dynamic Graph Learning Benchmarks.}
	While popular general graph learning benchmarks, such as \emph{Open Graph Benchmark} \citep{hu_open_2020}, also include link prediction tasks with time-based splits, benchmarks specifically targeting temporal graphs have emerged in recent years.
	Most notably, \emph{Temporal Graph Benchmark (TGB)} \citep{huang_temporal_2023} and \emph{BenchTemp} \citep{huang_benchtemp_2024} evaluate temporal graph learning models on node, link and graph classification tasks, showing that although strong results can be achieved on most datasets, there is no single solution across distinct datasets.
	\emph{TGB} was later extended to heterogeneous and knowledge graphs \citep{gastinger_tgb_2024}. \citet{yi_tgbseq_2025} identified limitations of temporal graph learning models in capturing sequential patterns in data, and established a benchmark focused on such data, while \citet{gravina_deep_2024} built separate benchmarks for spatio-temporal, discrete-time, and continuous-time graph datasets.
	The distinction between discrete- and continuous-time models is bridged by \citet{huang_utg_2025}, who present a unified framework and find that, despite some information loss, discrete-time models yield competitive results compared to continuous-time models, while being substantially faster at inference.
	
	\subheader{Limitations of Temporal Link Prediction Models.}
	Several works have identified limitations in link prediction evaluation and the predictive capabilities of temporal graph learning models.
	\citet{poursafaei_better_2022} found that strong performances in dynamic link prediction is often explained by task simplicity, as evidenced by their \emph{EdgeBank} baseline. EdgeBank predicts a positive edge at test time if and only if it was observed in training, yielding similar accuracy to state-of-the-art methods on several datasets. Thus, EdgeBank was also included in benchmarks such as \emph{TGB} \citep{huang_temporal_2023} and \emph{BenchTemp} \citep{huang_benchtemp_2024}.
	Reported performance can also be skewed by batch-based evaluation protocols \citep{lampert_link_2026}.
	\citet{rahman_rethinking_2025} demonstrated limitations of temporal graph learning models in learning temporal patterns, as perturbations such as repeating positive edges with slightly altered timestamps or shuffling timestamps among training edges often have minimal impact on performance.
	Additionally, \citet{cornell_power_2025} showed that simple heuristics favoring links to popular or recently active nodes can outperform state-of-the-art models on several datasets from \emph{TGB} and \emph{BenchTemp}.
	They further identified flaws in evaluation practices: commonly used rank-based metrics computed on sampled edge sets can yield results inconsistent with (computationally expensive) rankings on the full edge sets.
	In a broader context, \citet{bechler-speicher_position_2025} warned that poor benchmarking could lead to graph learning as a field losing relevance.

	\begin{table}[b]

\caption{\capheader{Statistics of empirical datasets.} 
\emph{Continuous Edges} denotes the number of edges in the original continuous-time datasets, \emph{Discrete Edges} the number of edges after timestamp discretization and removal of duplicate edges, and \emph{Unique Edges} the number of distinct edges irrespective of time.
}
\label{tab:datasets}
\medskip
\small
\centering
\begin{tabular}{lcccccc}
\toprule
\textbf{Dataset} & \textbf{Nodes} & \makecell{\textbf{Continuous}\\ \textbf{Edges}} & \makecell{\textbf{Discrete}\\ \textbf{Edges}} &  \makecell{\textbf{Unique}\\ \textbf{Edges}} & \makecell{\textbf{Discrete}\\ \textbf{Timesteps}} \\
\midrule
Enron & 184 & 125,235 & 10,472 & 3,125 &  45 (monthly) \\
Bitcoin-Alpha & 3,783 & 24,185 & 24,185 & 24,185 & 63 (monthly) \\
UCI & 1,899 & 59,835 & 26,628 & 20,296 & 29 (weekly) \\
Wikipedia & 9,277 & 157,474 & 65,085 & 18,257 & 745 (hourly)\\
\bottomrule
\end{tabular}
\end{table}

	\section{Experimental Framework}
	
	In Section \ref{sec:results}, for each studied characteristic we systematically introduce the approach behind our evaluation, and follow it immediately with the findings. 
	Within this section, we discuss experimental decisions that impact multiple characteristic experiments. 
	Further, we provide the code used for our experiments on \url{https://github.com/dess-mannheim/tgl}.

	\subheader{Datasets.}
	In our experiments, we selected datasets specific to each characteristic, using both empirical and synthetic data.
	As empirical datasets, we chose the \emph{Bitcoin-Alpha} \citep{kumar_edge_2016}, \emph{Enron} \citep{shetty_enron_2004},  \emph{UCI} \citep{panzarasa_patterns_2009} and \emph{Wikipedia} \citep{kumar_predicting_2019} datasets, due to both their popularity in related work and computationally reasonable size in light of the extensiveness of our experiments.
	For these datasets, we follow \citet{huang_utg_2025} for discretizing the timestamps, going from UNIX timestamps to monthly, weekly or hourly granularity--see also Table \ref{tab:datasets} where we summarize key statistics of these datasets. 
	The impact of this discretization is explored as the first characteristic under temporal granularity in our experiments (see Section \ref{sec:general_feats}). 
	Additional details on the empirical datasets are provided in Appendix \ref{ap:datasets}.
	For the temporal granularity, direction and density characteristics, the empirical datasets were split into training, validation, and test data based on these discretized time steps, with the first 70\% of times being used as training data, and the remaining 30\% evenly split into validation and test data. 
	For the other characteristics, we designed synthetic graphs with direct correspondence to the studied characteristic and present these in the relevant results section.

	\subheader{Models and Training Setup.} 
	In our framework, we consider DyGFormer \citep{yu_better_2024}, GraphMixer \citep{cong_we_2023}, DyRep \citep{trivedi_dyrep_2019}, JODIE \citep{kumar_predicting_2019}, TGN \citep{rossi_temporal_2020}, TCL \citep{wang_tcl_2021}, TGAT \citep{xu_inductive_2020} and CAWN \citep{wang_inductive_2021}, 
	using the implementations from \texttt{DyGLib} \citep{yu_better_2024}.
	We give an overview of these methods with regard to their architectural design choices in \Cref{tab:model_categorization}.
	As hyperparameters, we used fixed values that gave effective performance on the empirical datasets. The values are set out in Appendix \ref{ap:hyperparameters}, together with details of limited hyperparameter tuning in Appendix \ref{ap:alt_hp}.
	Unless noted otherwise, models were trained with an early stopping criterion based on accuracy on validation data. Where validation data was not logically related to the training data, the training loss was used instead. We trained models ten times on each dataset, using different training seeds, or five times for each of five generation seeds for synthetic datasets.  When not using the DyGLib benchmark evaluation setup, testing is based on predictions for all possible edges at test time.

	\begin{table}[t!]
\centering
\footnotesize
\begin{tabular}{
l
@{\hspace{0.75em}}c
@{\hspace{0.75em}}c
@{\hspace{0.75em}}c
@{\hspace{0.75em}}c
c
c
}
\toprule
& \multicolumn{4}{c}{Architecture}
& \multicolumn{2}{c}{Temporal information} \\
\cmidrule(lr){2-5}
\cmidrule(lr){6-7}
& Memory
& \makecell[c]{Graph\\ Attention}
& Transformer
& Mixer
& \makecell[c]{Time \\ Encoding}
& History used for embedding \\
\midrule
\textbf{CAWN}       &        &             & \checkmark &        & Learned & Temporal walks \& node-appearance encoding \\
\textbf{DyGFormer}  &        &             & \checkmark &        & Learned & First-hop interaction histories \\
\textbf{DyRep}      & \checkmark & \checkmark &        &        & Learned & Recursive temporal neighborhoods \\
\textbf{GraphMixer} &        &             &        & \checkmark & Fixed   & Recent neighbor interactions \\
\textbf{JODIE}      & \checkmark &             &        &        & Learned & Node memory only \\
\textbf{TCL}        &        &             & \checkmark &        & Learned & Temporal neighbor sequences \\
\textbf{TGAT}       &        & \checkmark      &        &        & Learned & Recursive temporal neighborhoods \\
\textbf{TGN}        & \checkmark & \checkmark  &        &        & Learned & Recursive temporal neighborhoods \\
\bottomrule
\end{tabular}
\medskip
\caption{\capheader{Categorization of TGL Models.} We categorize the models under study in terms of core architectural building blocks they deploy, and the way they process temporal information. 
This categorization is based on the implementation of the models provided by \texttt{DyGLib}~\citep{yu_better_2024}.
Here, \emph{Graph Attention} denotes multi-head attention used for temporal neighborhood aggregation over sampled historical neighbors. 
By contrast, \emph{Transformer} denotes architectures that apply Transformer encoder blocks~\citep{vaswani_attention_2017} to graph-derived token sequences, such as interaction histories, temporal walks, or patches. 
\emph{Memory} indicates that the model maintains node-level memory states, while \emph{Mixer} indicates an MLP-Mixer~\citep{tolstikhin_mlpmixer_2021} as the main building block.
}
\label{tab:model_categorization}
\end{table}

	\section{Results}\label{sec:results}
	
	In the following, we provide in-depth descriptions of our experiments for each characteristic, along with the corresponding findings. For each model, we give an assessment of whether they learn a characteristic (\cmark), learn it to a limited degree (\tmark), or do not learn it (\xmark). Results are also summarized in Table \ref{tab:results_summary}.

	\subsection{General Graph Features}\label{sec:general_feats}
\begin{figure}[t!]
		\centering
		\includegraphics[width=\textwidth]{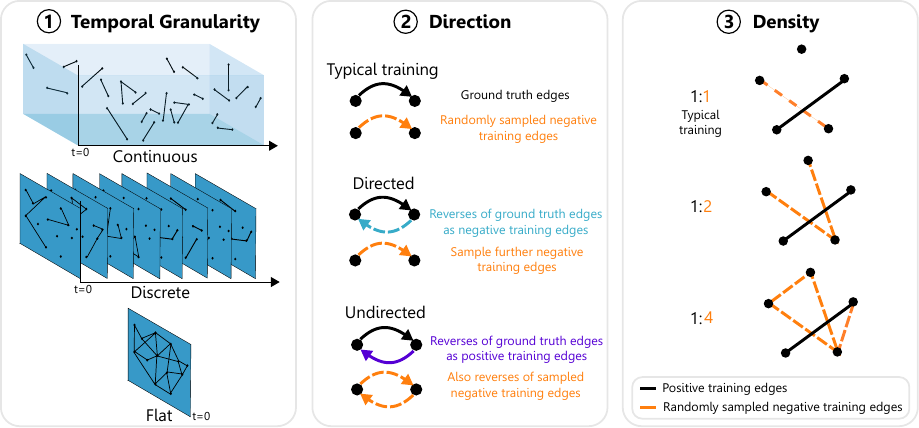}
		\caption{\capheader{General Graph Features.} 
			To evaluate whether temporal graph learning (TGL) models capture \emph{temporal granularity}, we vary the granularity of the edge timestamps in the data they are trained on, and test how performance on test data is affected.
			Specifically, we train the models using the original continuous timestamps, a discrete-time variant, and a flat variant where all training edges are allocated the timestamp $t=0$. 
			To assess whether TGL models capture \emph{direction} of edges, we compare the predicted probabilities of positive test edges and their non-existing reverse edges under typical training conditions. We then contrast these differences with those obtained under two modified training settings: in the directed setting, the reverses of the ground truth edges are included as explicit negative training edges, and in the undirected setting, the reverses of all ground truth edges and sampled negative edges are also included in the training data.
			Finally, we investigate whether TGL models capture the \emph{density} of a graph by evaluating whether their predictions reproduce either the density observed during training or the ground-truth density of the test graph. For that purpose, we adapt training data by varying the ratio between positive and negative training edges, starting from the typical training ratio of 1:1 and successively doubling the number of negative training examples toward the much lower ground-truth ratio. 
		}
		\label{fig:diagram_tg_dir_dens}
	\end{figure}

	We begin by presenting our experiments regarding \emph{temporal granularity}, \emph{direction}, and \emph{density}, which we also illustrate in \Cref{fig:diagram_tg_dir_dens}.

	\subsubsection*{\circled{1} Temporal Granularity}
	The timestamps of edges in continuous-time datasets are typically highly granular, often down to the minute or even second. This granularity is also typically assumed for link prediction at inference time, suggesting that predictions for very exact points in time can be made.
	To challenge this assumption, we investigate the effects of both using more discrete time steps and essentially removing temporal information altogether.
	
	\subheader{Approach.}
	For the three empirical datasets, we consider the following variants on which models are trained and tested:
	
	\begin{enumerate}
		\item \emph{Continuous:} the original datasets with UNIX timestamps.
		\item \emph{Discrete:} the discrete-time variants of the datasets, following \citet{huang_utg_2025}.
		\item \emph{Flat:} dataset variants with timestamps of all training edges set to 0, validation edges to 1, and test edges to 2.
	\end{enumerate}
	
	\noindent
	These variants are also illustrated in \Cref{fig:diagram_tg_dir_dens}, panel \circled{1}.
	We use training, validation and test splits based on the discrete data, such that no edge moves between groups.
	This avoids single time steps being separated into different splits. 
	We then evaluate performance on the test data, looking at whether benchmark performance deteriorates with coarser timestamps.
	Unless stated otherwise, discretized and flat variants retain duplicate edge events induced by timestamp coarsening. We ablate the effect of removing such duplicates separately in Appendix \ref{ap:results}, where we show that de-duplication can also strongly impact results, though the exact effects tend to be dataset-dependent.

	\subheader{Findings.} Table \ref{tab:time_granularity} shows the results across models and datasets for varying granularity.
	Flattening timestamps generally harms performance, often to a severe degree.
	This is a strong indicator that temporal information relevant for the link prediction is present and that the models use this information to produce improved predictions.
	Nevertheless, DyRep, JODIE, and TGN frequently achieve performance close to that of the best-performing variant even on flattened data, suggesting a weaker dependence on temporal information than other models.
	
	The effect of increasing temporal granularity beyond the discrete setting is considerably less consistent across datasets. While Enron is generally best learned with continuous timestamps, Bitcoin-Alpha often exhibits the opposite trend, with several models achieving better performance after discretization. On UCI and Wikipedia, differences between continuous and discrete timestamps are typically small. Consequently, the benefit of highly granular timestamps appears strongly dataset-dependent and substantially weaker than the benefit of preserving temporal information across multiple time steps.
	
	Overall, the only model that consistently benefits from continuous timestamps appears to be GraphMixer (\cmark). CAWN, DyGFormer, and TCL exhibit dataset-dependent benefits from continuous timestamps, performing best on the continuous Enron data, while on Bitcoin-Alpha, best performance is achieved on the discrete data. 
	DyRep, JODIE, and TGN achieve comparatively strong performance even when timestamps are flattened entirely, indicating that they rely heavily on signals other than temporal granularity. Nevertheless, these models typically still benefit from discrete over flattened timestamps, suggesting that temporal information contributes to their predictions, albeit to a lesser degree than for other models. By contrast, TGAT consistently performs better with discrete compared to continuous timestamps, yet suffers substantial performance degradation when timestamps are flattened. We therefore conclude that all remaining models capture temporal granularity only to a limited degree (\tmark).

	\begingroup
\setlength{\tabcolsep}{2.1pt}

\begin{table*}[t!]
\centering

\begin{tabular}{@{}l lll G lll G lll G lll@{}}
\toprule
&
\multicolumn{3}{c}{Bitcoin-Alpha}
&
&
\multicolumn{3}{c}{Enron}
&
&
\multicolumn{3}{c}{UCI}
&
&
\multicolumn{3}{c}{Wikipedia} \\
\cmidrule(lr){2-4}
\cmidrule(lr){6-8}
\cmidrule(lr){10-12}
\cmidrule(lr){14-16}
&
\multicolumn{1}{c}{Cont.} &
\multicolumn{1}{c}{Disc.} &
\multicolumn{1}{c}{Flat}
&
&
\multicolumn{1}{c}{Cont.} &
\multicolumn{1}{c}{Disc.} &
\multicolumn{1}{c}{Flat}
&
&
\multicolumn{1}{c}{Cont.} &
\multicolumn{1}{c}{Disc.} &
\multicolumn{1}{c}{Flat}
&
&
\multicolumn{1}{c}{Cont.} &
\multicolumn{1}{c}{Disc.} &
\multicolumn{1}{c}{Flat} \\
\midrule
\textbf{CAWN} & 0.780 & 0.984\sig & 0.728\sig &  & 0.959 & 0.903\sig & 0.630\sig &  & 0.968 & 0.971 & 0.671\sig &  & 0.989 & 0.988\sig & 0.625\sig \\
\textbf{DyGFormer} & 0.862 & 0.989\sig & 0.724\sig &  & 0.952 & 0.922\sig & 0.639\sig &  & 0.964 & 0.972\sig & 0.463\sig &  & 0.988 & 0.986\sig & 0.507\sig \\
\textbf{DyRep} & 0.893 & 0.905 & 0.941 &  & 0.891 & 0.830\sig & 0.810 &  & 0.925 & 0.935 & 0.916\sig &  & 0.965 & 0.965 & 0.958\sig \\
\textbf{GraphMixer} & 0.997 & 0.993\sig & 0.439\sig &  & 0.951 & 0.823\sig & 0.459\sig &  & 0.983 & 0.944\sig & 0.518\sig &  & 0.975 & 0.955\sig & 0.619\sig \\
\textbf{JODIE} & 0.999 & 0.999\sig & 0.961\sig &  & 0.936 & 0.911\sig & 0.853\sig &  & 0.958 & 0.964\sig & 0.941\sig &  & 0.969 & 0.971\sig & 0.954\sig \\
\textbf{TCL} & 0.880 & 0.991\sig & 0.668\sig &  & 0.842 & 0.832 & 0.554\sig &  & 0.956 & 0.969\sig & 0.677\sig &  & 0.971 & 0.960\sig & 0.625\sig \\
\textbf{TGAT} & 0.838 & 1.000\sig & 0.582\sig &  & 0.756 & 0.844\sig & 0.488\sig &  & 0.869 & 0.925\sig & 0.491\sig &  & 0.968 & 0.966 & 0.522\sig \\
\textbf{TGN} & 0.995 & 0.999\sig & 0.955\sig &  & 0.908 & 0.912 & 0.904 &  & 0.982 & 0.971\sig & 0.915\sig &  & 0.985 & 0.979\sig & 0.967\sig \\

\bottomrule
\end{tabular}

\caption{\capheader{Temporal Granularity}: \emph{impact of granularity of timestamps on performance.} 
We show average ROC-AUC of graph learning models on benchmark test sets for varying time granularities. \emph{Cont.} indicates models were trained on the original data with UNIX timestamps, \emph{Disc.} indicates data with discretized timestamps, and \emph{Flat} denotes training data where all timestamps were set to 1. 
A $\sig$ indicates that the corresponding AUC score differs from the score achieved at the next higher granularity to a degree that is statistically significant with respect to a permutation test.
We observe that only GraphMixer consistently improves its performance with more granular timestamps (\cmark).
All remaining models also generally strongly deteriorate in performance when timestamps are flattened. 
Yet, they do not do not in general appear to benefit from continuous over discrete timestamps (\tmark).
}
\label{tab:time_granularity}
\end{table*}

\endgroup
	
	\subheader{Implications.}
	Our results suggest that highly granular timestamps do not universally provide more useful information for temporal link prediction. While \mixer{} consistently benefits from continuous timestamps, the remaining models often exhibit only marginal gains or even deteriorate when moving from discrete to continuous time, with Bitcoin-Alpha providing the clearest example. This indicates that additional timestamp precision can provide useful signal in some graphs, but may also introduce noise or fine-scale variation that current models do not exploit effectively. Even when timestamps are recorded at high resolution, the finest available resolution may not be the most meaningful prediction target. In many practical applications, predicting whether an interaction occurs within a broader time window may be more sensible than predicting the exact second at which it occurs. 
	Relatedly, \citet{rahman_rethinking_2025} also showed that temporally distorting test data often has limited impact on temporal link prediction performance, similarly suggesting that current TGL models do not consistently exploit fine-grained temporal information.
	
	Interestingly, the three models that retain particularly strong performance on flattened timestamps---\dyrep{}, \jodie{}, and \tgn{}---all employ node memories (cf. \Cref{tab:model_categorization}), suggesting that maintaining persistent node states may reduce reliance on precise temporal information.
	By contrast, GraphMixer is the only evaluated model that consistently benefits from continuous timestamps, potentially owing to its fixed temporal encoding, which outperformed learned alternatives in the original GraphMixer study \citep{cong_we_2023}.
	While this observation remains qualitative, it provides a promising direction for future studies investigating how architectural design influences the utilization of temporal information.

	\subsubsection*{\circled{2} Direction}
	In many empirical networks, edge direction is important: such as when encoding a signal sent from a source to a destination node. In graph learning models, direction is typically implied by node order, though directionality is not always explicitly assumed. Consequently, during inference, predicted edge probabilities are not necessarily symmetric with respect to node order.

	\subheader{Approach.} To identify whether temporal graph learning models indeed learn the direction of edges, we train them on the discretized empirical datasets under the following different training settings, which are also illustrated in \Cref{fig:diagram_tg_dir_dens}, panel \circled{2}:
	\begin{enumerate}
		\item \emph{Typical Training}: we use the original positive training edges, and sample negative edges uniformly at random from all edge pairs that do not have an edge.
		\item \emph{Directed:} for each positive training edge $(u,v,t)$, we add the reverse edge $(v,u,t)$ as \emph{negative} edge to the training data, provided it does not appear as a positive edge. We additionally sample negative edges uniformly at random from all node pairs that do not have an edge.
		\item \emph{Undirected}: we make both positive and negative edges bidirectional by adding their reverse edges to the training data, i.e., if $(u,v,t)$ appears as positive or negative edge, then so must $(v,u,t)$.
	\end{enumerate}
	
	\noindent
	Through this, we aim to explore the degree to which predictions of positive edges $(u,v, t)$ and their negative reverse edges $(v,u,t)$ differ.
	If models learn directionality, a difference should be noticeable in the typical training setting, strengthened further if their reverse edges are explicitly negative.
	Conversely, explicitly training on both directions should narrow the gap between the probability scores of edges and their reverses, ideally nullifying this difference.
	In all cases, we aimed to maintain (roughly) a 1:1 ratio of positive edges to randomly sampled negative edges, excluding the negative reverse edges in the \emph{directed} setting. In the case that there are $n$ positive edges, we would normally use $n$ sampled negative edges. To ensure at least this number of sampled negative edges is also used, the models are trained with $2n$ negative edges for this variant, where up to $n$ may be the reverses of positive edges.
	
	The evaluation considers positive test edges $(u,v,t)$ from the regular discretized test data and their corresponding reverse edges $(v,u,t)$, where the reverse edge is not observed at the same time step. For each such pair, we compute the absolute difference between the predicted probabilities of $(u,v,t)$ and $(v,u,t)$. We then compare the distribution of these differences across the three training settings.
	
	\begin{figure}[t!]
		\centering
		\includegraphics[width=\textwidth]{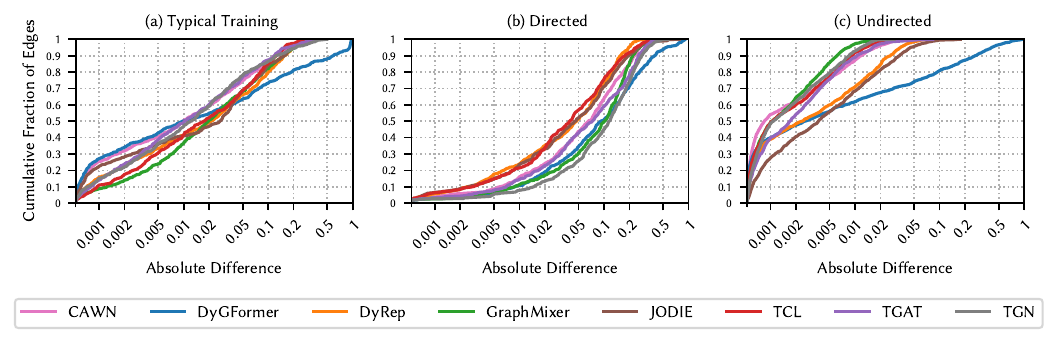}
		\caption{\capheader{Direction}: \emph{ability of graph learning models to distinguish edge directions.} 
			For each positive edge $(u,v)$ in the UCI test data, we compute the absolute difference between the probability predicted for $(u,v)$ and the probability predicted for its non-existent reverse edge $(v,u)$. 
			Plot (a) shows the cumulative distribution of these differences under the \emph{typical training} setting, where models are trained on the original training edges.
			Plot (b) shows the corresponding distribution under the \emph{directed} training setting, where reverse edges of positive training edges are provided as negative training samples.
			Plot (c) shows the corresponding distribution under the \emph{undirected} training setting, where both positive and negative training edges are provided in both directions.
			We observe under typical training, for most models, for roughly 50\% of all edges, the predicted probabilities of positive edges and their reverses differ by less than 0.02. 
			Providing reverse edges as explicit negatives increases these differences, reducing the share of such edge pairs with differences below $0.02$ to roughly $30\%$. 
			Conversely, training bidirectional edges increases symmetry in predicted edge probabilities, with 90\% of all edges having a difference less than 0.01 for many models and only \dygformer{} yielding high differences.
			Overall, this indicates limited ability of models to learn direction (\tmark). 
		}
		\label{fig:direction_uci}
	\end{figure}

	\subheader{Findings.}
	We depict results for UCI in \Cref{fig:direction_uci}, with largely consistent results on other datasets included in Appendix \ref{ap:results}.
	For the original training data with implied directionality, we observe that most models assign highly similar probabilities to true edges $(u,v,t)$ and the negative reverse edge $(v,u,t)$ at testing time.
	For roughly half of the edges, the probability scores differ by less than 0.02, and, except for \dygformer{}, the differences are bigger than $0.1$ for less than 20\% of the positive edges.
	When training with explicitly directed edges, differences in probability scores between reverse edge pairs tend to increase a moderate amount.
	Conversely, when always including both directions of an edge in the training data, we obtain nearly symmetric predictions across all positive edges. The only exception is \dygformer{}, which still yields differences bigger than 0.1 in the probability scores.
	Overall, we find that models do not clearly distinguish edge direction under standard training, but that explicitly including reverse edges in the training signal can steer predictions toward either more directed or more symmetric behavior. We therefore conclude that models capture edge direction only to a limited degree (\tmark).
	
	This pattern is further corroborated in Appendix \ref{ap:results}, Table \ref{tab:direction}: across datasets, non-existent reverse edges receive scores much closer to observed directed edges than to other negative edges under typical training, with the directed training setting reducing this gap only partially.

	\subheader{Implications.}
	These results suggest that current TGL models have no intrinsic notion of edge direction, which is unsurprising since the input representation itself does not convey the semantic distinction between an edge and its reverse, instead simply presenting ordered node pairs and supervision labels.
	This yields a takeaway for practical applications: when edge direction is semantically important, the training data should be designed according to our \emph{directed} or \emph{undirected} training setting, rather than relying on the model to infer 
	automatically.

	\begin{figure}[t]
		\includegraphics[width=\textwidth]{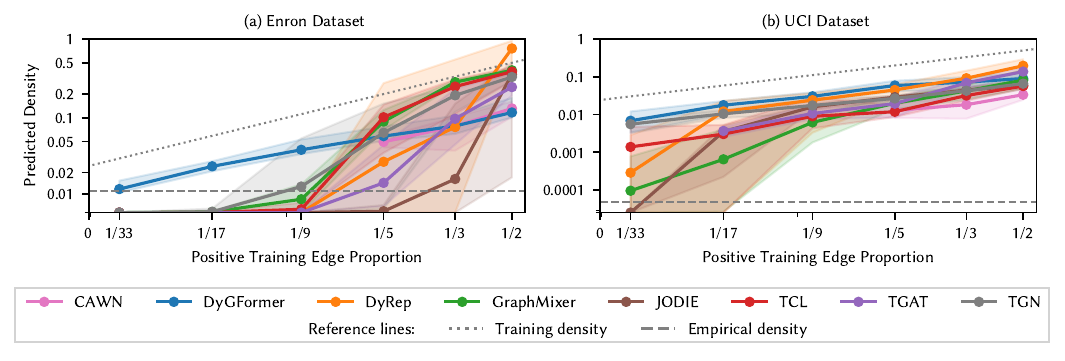}
		\caption{\capheader{Density:} \emph{ability of models to replicate true density of networks.} We trained on the same set of positive training edges, but varied the negative sampling ratio. We depict the density resulting from predicting on all potential edges. Predicted density is generally much lower than the density seen during training. True density also appears hard to approximate, as models seem prone to predicting no edges when seeing larger numbers of negative edges. Thus, models do not appear to learn density (\xmark).
		}
		\label{fig:density}
	\end{figure}

	\subsubsection*{\circled{3} Density}
	Link prediction benchmarks typically evaluate the performance of graph learning models on only a small subset of all possible links, often because a full evaluation would be extremely computationally expensive. 
	As a result, however, it remains unclear whether models can grasp the density of a network, one of the most fundamental graph characteristics.

	\subheader{Approach.}
	To evaluate whether graph learning models learn the density of a graph, we train the models on the discretized empirical datasets and vary the number of negative training edges during training: starting from a 1:1 ratio of training to test edges, and doubling the negative edges successively, as illustrated in \Cref{fig:diagram_tg_dir_dens}, panel \circled{3}. For CAWN, this was limited to a maximum of 1:4, since more training data would exceed 80GB VRAM. Evaluation is done by making a prediction for all possible edges at test time and recording the resulting density. Intuitively, a model should either naturally mimic the ground truth density, or the predicted density should be proportional to the ratio of positive to negative training edges, which is the only explicit signal for density during training.

	\subheader{Findings.}
	In \Cref{fig:density}, we observe that the predicted density is consistently much lower than the training ratio of positive to negative edges across all evaluated models. This indicates that models do not simply reproduce the class distribution observed during training. However, for commonly used training ratios, the predicted density still exceeds the true graph density by several orders of magnitude. Increasing the number of negative training samples slowly reduces the predicted density, before models eventually collapse to predicting nearly all edges as negative. When doing this, a model may occasionally get close to the true density, such as \dygformer{} on the Enron dataset at ratio of 1:32 positive to negative edges.
	Overall, these results indicate that the evaluated models do not capture graph density (\xmark).
	
	
	
	\subheader{Implications.}
	A possible explanation for the inability of current TGL models to recover realistic graph densities is that the models are never exposed to the total number of possible edges in the graph, instead only receiving observed edges and sampled negative edge examples. Consequently, graph density is not an explicit quantity in the learning problem.
	
	From a practical perspective, this limitation may be of little concern for ranking-based applications such as recommender systems, where relative ordering of candidate edges is more important than the absolute number of predicted interactions. However, in settings where accurate prediction of future interactions is crucial, recalibration of the output probability scores may be needed.
	
	Further, our experiments suggest that simply increasing the number of negative training examples is unlikely to provide a practical solution to this limitation.
	One option would be to train models with multiple positive-to-negative ratios and select the one that best approximates a realistic graph density, but this would require several costly training runs. Alternatively, a higher ratio could be used without such tuning, however our results suggest that overly sparse training distributions can instead drive models toward predicting nearly all edges as negative.

	\begin{figure}[t]
		\centering
		\includegraphics[width=\textwidth]{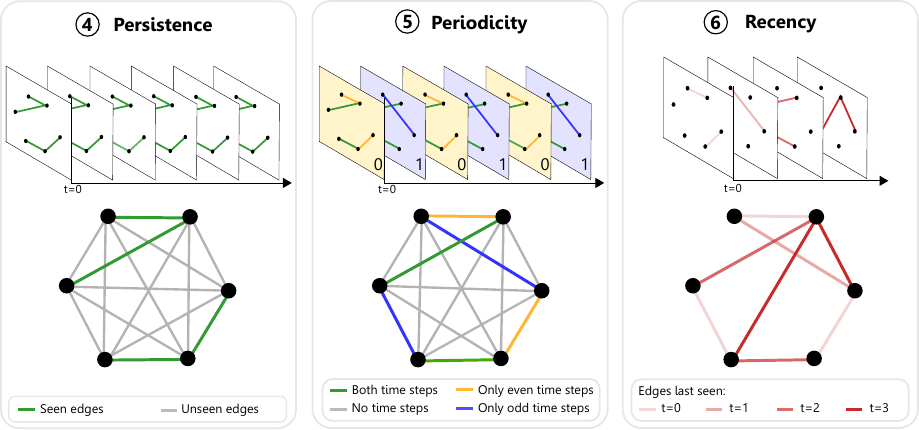}
		\caption{\capheader{Temporal Patterns.} 
			To evaluate whether temporal graph learning (TGL) models capture \emph{persistence}, we train them on graphs in which the same snapshot is repeated at all time steps. A model should then be able to distinguish seen edges from unseen edges.
			To determine whether TGL models capture \emph{periodicity}, we train them on graphs in which the same pair of snapshots is repeated throughout all time steps. We then evaluate predictions at odd and even time steps, testing whether models distinguish edges that appear at all time steps, only at odd time steps, only at even time steps, or at no time step.
			Finally, we evaluate whether TGL models capture \emph{recency} by partitioning the edges of a graph across successive time steps such that each edge is observed only once during training.
			We then test whether edges seen more recently receive higher predicted probabilities than edges seen earlier.
		}
		\label{fig:diagram_pers_peri_rec}
	\end{figure}

	\subsection{Temporal Patterns}
	Next, we evaluate the ability of temporal graph learning models to learn \emph{persistence}, \emph{periodicity}, and \emph{recency}, as seen in \Cref{fig:diagram_pers_peri_rec}.

	\begin{figure}[t]
		\centering
		\includegraphics[]{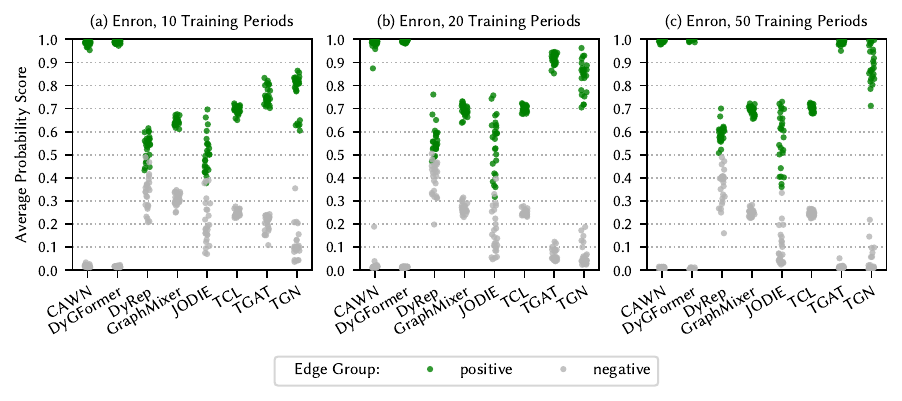}
		\caption{\capheader{Persistence:} \emph{ability of models to learn persistent graphs.}
			We trained the graph learning models on fixed snapshots from the Enron dataset, which were repeated throughout training, and depict average probability scores resulting from each model when predicting positive and negative edges of these snapshots. 
			Only \cawn{} and \dygformer{} appear to reproduce the fixed graph pattern early on with high confidence (\cmark), while all other models generally discriminate positive and negative edges after a sufficient amount of time steps (\tmark). 
		}
		\label{fig:persistence_enron}
	\end{figure}

	\subsubsection*{\circled{4} Persistence}
	Persistent link patterns often occur in empirical data, and, intuitively, these should be trivial to learn.
	Thus, we investigate whether temporal graph learning models can reproduce such patterns, considering the extreme case of fully constant graphs.
	
	\subheader{Approach.} 
	We create fully persistent graphs by sampling a single snapshot from an empirical dataset, and repeating this snapshot $T$ times, where we consider $T\in\{10,20,50, 100\}$.
	The number of time steps $T$ is varied so that we can evaluate whether individual models are slower at picking up the persistent pattern than others. 
	All nodes that are not present at the sampled snapshot are excluded from the network, and during training, negative samples are redrawn at every time step.
	For evaluation, we compare between aggregate predicted probabilities of \emph{seen edges} that exist in the snapshot and the corresponding aggregate probabilities of all \emph{unseen edges} that could exist between the given nodes, as illustrated in \Cref{fig:diagram_pers_peri_rec}, panel \circled{4}.

	\subheader{Findings.} In \Cref{fig:persistence_uci}, we present aggregated confidence scores of positive and negative edges on the Enron dataset, collected after training the persistent snapshot for 10, 20 and 50 time steps. 
	Overall, all models eventually learn to assign higher probabilities to seen edges compared to unseen edges, indicating that they capture the persistent graph to some extent. 
	However, substantial differences exist in how quickly and how consistently this separation is achieved. \cawn{} and \dygformer{} already distinguish positive and negative edges with near-perfect confidence after 10 training periods and consistently retain this behavior across datasets (\cmark). 
	
	\mixer{}, \tcl{}, \tgat{}, and \tgn{} also establish a clear separation after only a few training periods, whereas \dyrep{} and \jodie{} require longer until full separation is achieved, often still only by small margins.
	These trends are consistent across other datasets, although models sometimes require longer until the pattern is learned.
	Yet, despite this apparent separation, the corresponding balanced accuracies remain surprisingly low for these six models, even after 50 training periods (cf. Appendix \ref{ap:results}, \Cref{fig:persistence_accuracies}). This indicates that, although the models learn to distinguish persistent from unseen edges in aggregate, they fail to reproduce the constant graph with consistently high accuracy. We therefore conclude that these models only capture \emph{persistence} to a limited degree (\tmark).
	
	\subheader{Implications.}
	Persistence appears to be a characteristic that current TGL models can generally learn, with the primary differences lying in how quickly and consistently it is captured.
	One possible explanation for the particularly strong performance of \cawn{} and \dygformer{} is that both architectures explicitly model interaction histories, although in different ways (cf. \Cref{tab:model_categorization}). While \dygformer{} processes extended first-hop interaction histories, \cawn{} explicitly encodes node appearances within historical walks, thereby emphasizing recurring interaction partners. 
	The remaining models also incorporate historical information, but represent it through sampled temporal neighborhoods or compressed node memories, which may be less well suited to capturing repeated interaction patterns.

	\begin{figure}[t]
		\centering
		\includegraphics[]{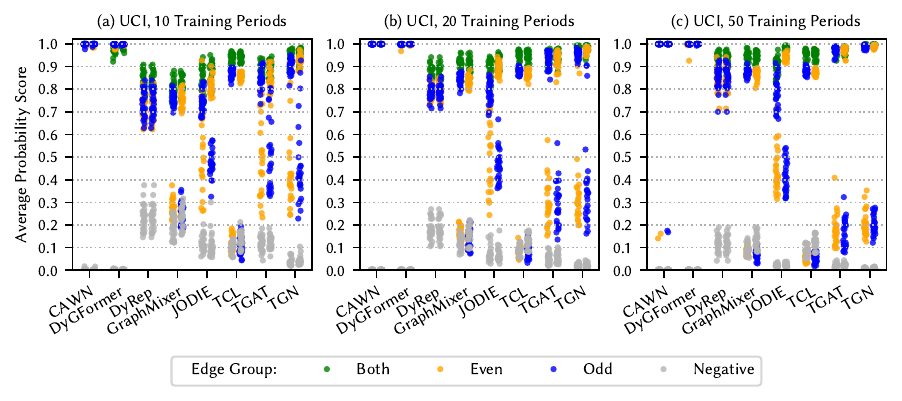}
		\caption{\capheader{Periodicity:} \emph{ability of models to learn periodically repeated edges.} We selected pairs of consecutive snapshots from the UCI dataset, and tested whether the temporal graph learning models could reproduce a consistent pattern of two oscillating snapshots. We depict average predicted probabilities when testing at odd (left) and even (right) time steps, colors correspond to predictions on edges present at odd, even, both, or neither time step.
			Overall, most models appear eventually to reproduce the periodic pattern (\cmark), although \jodie{} takes longer to achieve separation and still appears to predict a few false positives (\tmark).
			Only \cawn{}, \dygformer{} assign high confidence to all edges which ever appear as positive, but distinguish odd and even edges on other datasets (\tmark), see Appendix \ref{ap:results}.
			\dyrep{} assigns higher probabilities for edges that occur more frequently, but does not appear to distinguish odd and even time steps (\xmark).
		}
		\label{fig:periodicity_uci}
	\end{figure}
	
	\subsubsection*{\circled{5} Periodicity} 
	Going beyond completely constant networks, we now move to networks with repeating patterns. 
	Temporal graph learning models should, intuitively, also be able to identify periodic patterns, i.e., networks which repeat their edges every $k$ time steps.

	\subheader{Approach.}
	We create graphs with period length $k\in\{2,5\}$ from empirical networks by sampling an initial time step at which the current and subsequent $k-1$ snapshots are taken.
	We then repeat this sequence of $k$ snapshots $m$ times, with $m\in\{10,20,50\}$ such that in total, we obtain graphs with $T=mk$ time steps. Again, we vary the number of times the pattern is seen to identify whether models learn to identify the given pattern over time.
	From the resulting graphs with $k=2$, each pair of continuously repeated snapshots yields four groups of edges, which are also illustrated in \Cref{fig:diagram_pers_peri_rec}, panel \circled{5}: (i) edges which appear at \emph{both time steps}, (ii) edges which appear \emph{only at odd time steps}, (iii) edges which appear \emph{only at even time steps}, and (iv) edges which never appear (\emph{no time steps}). 
	In our evaluation, we distinguish average predicted probabilities for these groups of edges both at odd and even test time steps.
	For $k=5$, the analysis is similar, see Appendix \ref{ap:results}.

	\subheader{Findings.}
	Results for periodicity in terms of aggregate confidence scores after training for 10, 20 and 50 time steps on the UCI dataset are depicted in \Cref{fig:periodicity_uci}. These are largely in line with the results for other datasets, period lengths, and the corresponding accuracy scores in Appendix \ref{ap:results}.
	We observe some differences compared to our experiments on persistence.
	
	Although \cawn{} and \dygformer{} perfectly captured persistent graphs, they fail to distinguish between odd and even time steps, assigning similarly high confidence to every edge that appears in the repeated snapshots. This behavior closely resembles that of the \emph{EdgeBank} baseline \citep{poursafaei_better_2022}, effectively memorizing which edges have been observed, without distinguishing the time steps at which they occur.
	Yet, on the Enron dataset (see Appendix \ref{ap:results}, \Cref{fig:periodicity_enron}), these methods appear to distinguish odd and even datasets, so we conclude that these methods have limited ability to learn periodicity (\tmark).
	However, \dyrep{} does not distinguish between odd and even time steps on any dataset, albeit with less extreme confidence scores (\xmark).
	By contrast, \mixer{} and \tcl{} distinguish odd and even time steps very well after only a few repetitions, while \tgat{} and \tgn{} gradually develop a similarly clear separation as training progresses (\cmark)
	Finally, \jodie{} consistently separates edges from different time steps, but tends to assign overly high probabilities to edges from the wrong time step (\tmark{}).
	
	\subheader{Implications.}
	The contrast between persistence and periodicity suggests that capturing recurring interactions and capturing their temporal regularity are distinct capabilities of current TGL models. \mixer{} illustrates this distinction particularly well: while it is not among the strongest models for persistence, it consistently performs well on both periodicity and temporal granularity. This may be due to its fixed time encoding, which is directly incorporated into the link encoder (cf. \Cref{tab:model_categorization}). 
	By contrast, the models that most consistently capture persistence, namely \cawn{} and \dygformer{}, may prioritize recurring interactions over the time at which they occur.

	\subsubsection*{\circled{6} Recency}
	For all deep learning models, it is often anticipated that models are more influenced by the latest data seen in training. This is even more of a consideration when the data itself contains a temporal element, as with dynamic graphs. Recent work has also shown that heuristics which emphasize predicting edges on more recently active nodes can perform very strongly on graph datasets \citep{cornell_power_2025}. Thus, we look to evaluate the extent to which model predictions are biased toward more recently seen data.
	
	\begin{figure}[t]
		\includegraphics[width=\textwidth]{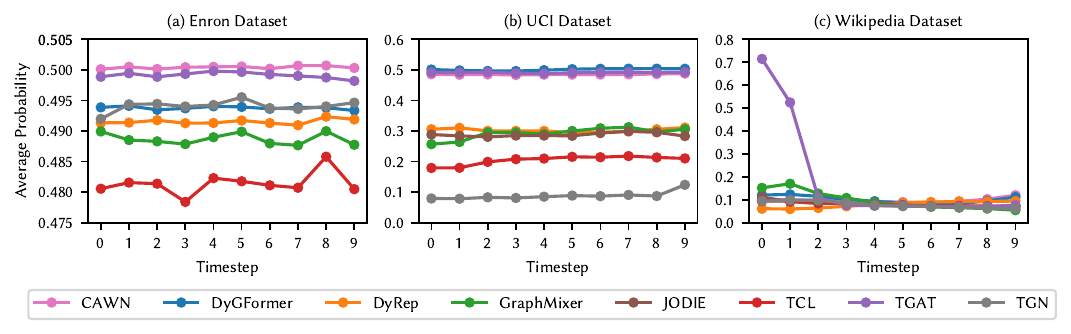}
		\caption{\capheader{Recency:} \emph{impact of time that an edge was last seen on its probability score at test time.} For 10 time steps, we sampled a random set of positive edges. These edge sets are disjoint over all time steps, and reflect the density at representative time steps in the original corresponding dataset. We show average predicted probability scores at time step $t=10$ for all positive edges seen during training, separated by the time step in which they were seen.
			Overall, we observe that there is no consistent trend regarding whether more recently (or earlier) edges have higher probability scores (\xmark).
			Instead, all edges appear to have very similar probability scores on average, with the exception of \tgat{} on the graphs relating to the Wikipedia dataset.
		}
		\label{fig:recency}
	\end{figure}

	\subheader{Approach.}
	To evaluate a model's ability to learn \emph{recency}, we create datasets of 10 time steps with disjoint sets of edges, and evaluate whether more recently seen edges are allocated a higher probability score than earlier edges, as illustrated in \Cref{fig:diagram_pers_peri_rec}, panel \circled{6}. 
	These datasets are related to three empirical datasets by taking the number of edges at a representative time step and sampling the same number of edges uniformly at random from all potential edges. This is done for each time step successively, without replacement. 
	In negative sampling, we exclusively sample edges which are not positive at any time step. For both positive and negative edges, only one direction of each edge is considered for sampling.
	Thus, edge probabilities cannot be biased by some positive edges also being chosen as negative examples at other time steps.
	No validation data is used, as there is no explicit connection between different time steps, and predictions are made at time $t=10$.

	\subheader{Findings.} 
	In \Cref{fig:recency}, we observe that, across all models and datasets, average probability scores of seen edges remain largely constant regardless of when edges were last observed. The only notable exception is \tgat{}, which appears biased toward earlier edges on the Wikipedia-based graphs. Since edges were sampled completely at random, there is no additional information on which models could base their predictions, resulting in high uncertainty in the predicted probability scores. Overall, we conclude that the evaluated models do not place predictive weight on the recency of edges (\xmark).

	\subheader{Implications.}
	The failure to capture recency suggests that temporal information alone does not imply a preference for recently observed interactions. 
	This is notable because heuristics based on predicting edges to recently active nodes often perform on a par with state-of-the-art TGL models \citep{cornell_power_2025}. Our findings therefore indicate potential room for improvement in current TGL models. In particular, introducing mechanisms or parameters that control whether a model places more weight on recent observations could improve performance on datasets where recency is predictive.

	\subsection{Mechanisms in Edge Formation}
	Finally, we show results on \emph{homophily} and \emph{preferential attachment}, illustrated in Figure \ref{fig:diagram_hom_pa}.

	\subsubsection*{\circled{7} Homophily}
	Homophily refers to the phenomenon that `birds of a feather flock together'. Here attention is paid to whether edges are between individuals from the same or different groups. Models should be able to learn the homophily of a graph since they receive the group information and all edge data. 
	
\begin{figure}[t!]
		\centering
		\includegraphics[width=0.66\textwidth]{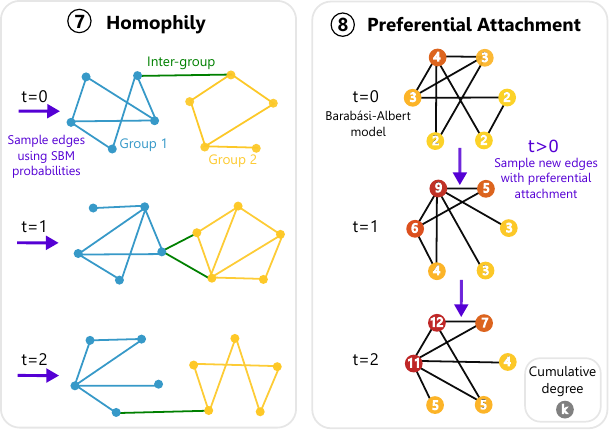}
		\caption{\capheader{Mechanisms in Edge Formation.} To assess whether TGL models capture \emph{homophily}, we train them on temporal graphs generated by repeatedly sampling snapshots from a stochastic block model with two groups of nodes, where intra-group edges are more likely than inter-group edges. We then test whether models assign higher probabilities to intra-group than inter-group edges. To determine whether TGL models capture \emph{preferential attachment}, we train them on temporal graphs initialized using a Barabási-Albert model, where edges at later time steps are sampled preferentially toward nodes that have already accumulated many edges. We then test whether unseen edges adjacent to high-degree nodes receive higher predicted probabilities than unseen edges adjacent to low-degree nodes.
		}
		\label{fig:diagram_hom_pa}
	\end{figure}

	\subheader{Approach.}
	To examine whether the models learn homophily, we use homophilic and heterophilic \emph{stochastic block models (SBMs)} \citep{holland_stochastic_1983} with 1000 nodes, split into two groups of 500. 
	For homophilic SBMs, we set the likelihood of inter-group links to $p=0.001$ and intra-group links to $p=0.005$, reversed for heterophilic SBMs.
	Based on these parameters, we created dynamic networks of $T=100$ time steps by resampling edges $T$ times, as illustrated in \Cref{fig:diagram_hom_pa}, panel \circled{7}.
	Intuitively, if a model learns homophily, it should mimic the edge formation of the homophilic SBM: placing higher likelihood on predicting intra-group rather than inter-group edges, having intra-group links for both groups being equally likely and showing opposite behavior for heterophilic SBMs.
	We test this assumption by aggregating the predicted probabilities of edges within and between groups.
	During training, the node groups are explicitly provided as one-hot-encoded node features.
	
	\subheader{Findings.} 
	Table \ref{tab:homophily_probs} presents the average predicted edge probabilities, as well as the share of edges predicted to be positive within all combinations of intra-group and inter-group links, in both the homophilic and the heterophilic settings.
	Except for \jodie{} and \tgn{} (\xmark), all models generally distinguish intra-group and intra-group edges in both the homophilic and the heterophilic setting.
	However, the predicted probabilities neither match the absolute SBM probabilities, nor reflect the intended ratio between majority and minority edge rates.
	Moreover, when considering binary predictions, most models amplify the stochastic group-level preference into near-deterministic predictions, predicting almost all edges of a group to appear or not.
	This extreme disparity is also reflected in the rankings of the top $k$ most likely edges (see Appendix \ref{ap:results}, Table \ref{tab:homophily_ranks}).
	The only exception is \dyrep{}, which assigns non-negligible probability to minority edge groups and predicts some edges as positive.
	Thus, we conclude that \dyrep{} captures homophily (\cmark), while all models other than \jodie{} and \tgn{} capture homophily to a limited degree (\tmark).

	\begin{table}[b!]
	\caption{\textbf{Homophily: }\emph{ability of models to reproduce homophily in edge formation.} We train graph learning models on stochastic block models with two groups (0 and 1) considering both homophilic and heterophilic graphs, with intra-group being five times more likely than inter-group links and vice versa.
		We depict the average condidence scores of all edges per group combination, and the share of all of edges per group which were predicted to exist.
		Except for \jodie{} and \tgn{} (\xmark), models generally distinguish intra-group and inter-group links (\tmark). 
		However, only \dyrep{} does not converge to fully binary deterministic predictions (\cmark).
	}
	\label{tab:homophily_probs}
	\medskip
	\centering
	\small
	\begin{tabular}{lrrrrrrrrrrrr}
		\toprule
		& \multicolumn{6}{c}{Homophilic SBM}
		& \multicolumn{6}{c}{Heterophilic SBM} \\
		\cmidrule(lr){2-7} \cmidrule(lr){8-13}
		& \multicolumn{3}{c}{\makecell[cc]{Avg. Edge\\Probability}}
		& \multicolumn{3}{c}{\makecell[cc]{Fraction of\\Predicted Edges}}
		& \multicolumn{3}{c}{\makecell[cc]{Avg. Edge\\Probability}}
		& \multicolumn{3}{c}{\makecell[cc]{Fraction of\\Predicted Edges}} \\
		\cmidrule(lr){2-4} \cmidrule(lr){5-7}
		\cmidrule(lr){8-10} \cmidrule(lr){11-13}
		Group
		& 0--0 & 0--1 & 1--1
		& 0--0 & 0--1 & 1--1
		& 0--0 & 0--1 & 1--1
		& 0--0 & 0--1 & 1--1 \\
		\midrule
		\textbf{CAWN} 		& 0.62 & 0.41 & 0.63 & 1.00 & 0.00 & 1.00 & 0.24 & 0.77 & 0.26 & 0.00 & 1.00 & 0.00 \\
		\textbf{DyGF}. 	& 0.62 & 0.41 & 0.62 & 1.00 & 0.00 & 1.00 & 0.24 & 0.76 & 0.26 & 0.00 & 1.00 & 0.00 \\
		\textbf{DyRep} 		& 0.61 & 0.33 & 0.55 & 0.74 & 0.35 & 0.62 & 0.41 & 0.67 & 0.24 & 0.10 & 0.86 & 0.09 \\
		\textbf{Mixer} 	& 0.61 & 0.42 & 0.62 & 1.00 & 0.04 & 1.00 & 0.24 & 0.77 & 0.24 & 0.00 & 1.00 & 0.00 \\
		\textbf{JODIE} 		& 0.81 & 0.17 & 0.40 & 0.91 & 0.00 & 0.40 & 0.71 & 0.73 & 0.02 & 0.91 & 0.88 & 0.00 \\
		\textbf{TCL} 		& 0.62 & 0.40 & 0.62 & 1.00 & 0.00 & 1.00 & 0.24 & 0.77 & 0.26 & 0.00 & 1.00 & 0.00 \\
		\textbf{TGAT} 		& 0.62 & 0.40 & 0.63 & 1.00 & 0.00 & 1.00 & 0.24 & 0.77 & 0.27 & 0.00 & 1.00 & 0.04 \\
		\textbf{TGN} 		& 0.30 & 0.21 & 0.17 & 0.33 & 0.16 & 0.14 & 0.70 & 0.15 & 0.43 & 0.88 & 0.14 & 0.46 \\
		\bottomrule
	\end{tabular}
\end{table}

	\subheader{Implications.}
	The given results suggest that most models learn which types of edges are more likely, but not how much more likely they are.
	Since the randomized nature of the SBM setting deliberately isolates group membership from all other predictive signals, this should not be read as direct evidence that the same behavior would occur in real-world graphs. Nevertheless, it suggests that current models may overstate homophilic or heterophilic tendencies when such group-level signals are predictive, potentially treating less likely edge categories as effectively absent.
	
	The memory-based models provide a partial exception to this binary behavior, although memory alone is not sufficient: \jodie{} and \tgn{} avoid fully deterministic predictions but do not consistently reflect the correct homophilic or heterophilic pattern, whereas \dyrep{} is the only model that both avoids binary collapse and preserves the expected group-level preferences.

	\begin{figure}[t!]
		\centering
		\includegraphics[]{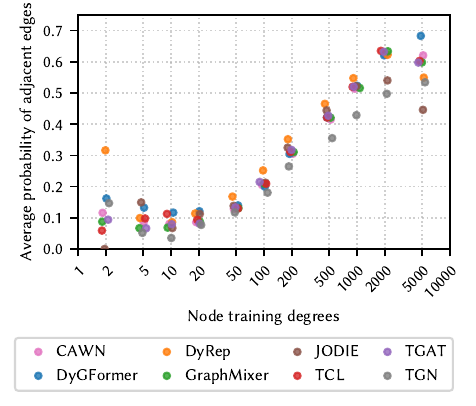}
		
		\caption{\capheader{Preferential Attachment:} \emph{ability of models to reproduce preferential attachment in edge formation.} We trained all models on dynamic preferential attachment graphs over 100 time steps.
			From the training graphs, we place the nodes into bins based on their logarithmic degree, and compute the average probability of all unseen edges adjacent to these nodes. 
			The x-axis denotes the lower bound of the corresponding bins.
			We observe that for all models, average probabilities continuously increase with exponential increase of node degree.
			Thus, all models learn preferential attachment (\cmark).
		}
		\label{fig:pa}
	\end{figure}

	\subsubsection*{\circled{8} Preferential Attachment}
	Preferential attachment refers to the phenomenon that high-degree nodes are more likely to receive new links. Given its frequent observation in empirical networks, it is of interest whether graph learning models learn this characteristic well.
	
	\subheader{Approach.}
	We create graphs based on the Barabási-Albert (BA) model \citep{barabasi_emergence_1999} with 1000 nodes and 2000 edges at each time step, for 100 training and 21 validation time steps---we also consider a denser variant with $4000$ edges per time step.
	The initial time step, $t=0$, is generated with the standard BA model setting the number of edges per incoming node to $m=2$, to introduce variation in node degree.
	Subsequent time steps add edges by sampling two existing nodes proportional to their number of previous edges, as illustrated in \Cref{fig:diagram_hom_pa}, panel \circled{8}. The probability $p_i^t$ that node $i$ is selected at time $t$ is defined as $p_i^t=\frac{\sum_{s=0}^{t-1}k_i^t}{\sum_{s=0}^{t-1}\sum_j k_j^s}$ where $k_i^s$ is the degree of node $i$ at time $s$.
	Evaluation is done by examining the relationship between each node's degree and the average probabilities predicted for those of its potential edges that were never positive in training. If preferential attachment is learned, then the edges connecting to nodes of high degree should receive higher average probabilities.
	
	\subheader{Findings.}  Results are presented in \Cref{fig:pa}, these are also in line with those for the denser graph depicted in in Appendix \ref{ap:results}, \Cref{fig:pa_dense}. We observe that all models assign very low probabilities on average, around 0.1, to edges relating to low-degree nodes, and, with an exponential increase of node degree, the average probability rises continuously up to 0.6-0.7. 
	Most models have very similar average probabilities per bin,  with only \tgn{} tending to appear slightly lower than the rest. 
	In addition, we see some slight outliers at the first and last bins, which could be attributed to the sparsity of these bins.
	Overall, the observed pattern is very reminiscent of the power-law degree distribution of the Barabasi-Albert model, and we conclude that all models indeed learn preferential attachment (\cmark). 
	
	\subheader{Implications.}
	Preferential attachment is the clearest positive case in our study. This is relevant because popularity-related signals can be highly predictive in temporal link prediction, with recent work showing that simple popularity-based heuristics can perform competitively with state-of-the-art TGL models \citep{cornell_power_2025}. Thus, all evaluated models capture a characteristic that is often useful for temporal link prediction.

	\section{Discussion}
	
	Overall, our results, as summarized in \Cref{tab:results_summary}, reveal clear differences in the graph characteristics captured by current temporal graph learning models. All evaluated models consistently struggle to capture direction, density, and recency, whereas preferential attachment is consistently learned across these methods. The remaining characteristics---temporal granularity, persistence, periodicity, and homophily---differentiate substantially between individual models, resulting in distinct profiles of learned graph characteristics. 
	In the following, we discuss how these findings relate to model design, consider their broader implications for temporal graph learning, and outline the limitations of our study.
	
	\subheader{Model Design and Learned Characteristics.}
	Several of the characteristic experiments point to possible connections between specific design choices and the graph characteristics that models learn. For instance, \dyrep{}, \jodie{}, and \tgn{} remain comparatively robust when timestamps are flattened, suggesting that persistent node states can reduce reliance on precise temporal information. Similarly, the strong performance of \cawn{} and \dygformer{} on persistence may be related to their explicit modeling of interaction histories, while the strong performance of \mixer{} on both temporal granularity and periodicity may be related to its fixed time encoding in the link encoder.
	
	At the same time, these observations do not translate into a simple mapping between broad architectural categories and learned graph characteristics. Models with node memories differ substantially in their behavior, as illustrated by the contrast between \dyrep{}, \jodie{}, and \tgn{} on homophily. Similarly, the use of Transformer-based components alone does not determine a model's profile of learned characteristics: while \cawn{} and \dygformer{} behave similarly across many of our experiments, \tcl{} differs notably despite also relying on transformer-based sequence modeling. Thus, high-level architectural components alone do not determine which characteristics a model can capture. 
	The observed differences instead suggest that finer-grained design choices may shape which characteristics are learned. Experiments that isolate individual graph characteristics, as done in this study, can help reveal such differences, while controlled architectural ablations are needed to test their underlying causes directly.
	
	\begin{table}[t!]
\centering
\small
\setlength{\tabcolsep}{3pt} 

\renewcommand{\arraystretch}{1.35}
\renewcommand\cellgape{\Gape[1pt][1pt]}


\begin{tabular}{@{}lccccccccc@{}}
\toprule
 & \textbf{CAWN}
 & \makecell[ccc]{\textbf{DyG}-\\\textbf{Former}} 
 & \textbf{DyRep} 
 & \makecell[ccc]{\textbf{Graph-}\\\textbf{Mixer}} 
 & \textbf{JODIE} 
 & \textbf{TCL} 
 & \textbf{TGAT} 
 & \textbf{TGN}  \\
\midrule
Temporal Granularity    & \tmark & \tmark & \tmark & \cmark & \tmark & \tmark & \tmark & \tmark \\
Direction                               & \tmark & \tmark & \tmark & \tmark & \tmark & \tmark & \tmark & \tmark \\
Density                                 & \xmark & \xmark & \xmark & \xmark & \xmark & \xmark & \xmark & \xmark \\
Persistence                             & \cmark & \cmark & \tmark & \tmark & \tmark & \tmark & \tmark & \tmark \\
Periodicity                             & \tmark & \tmark & \xmark & \cmark & \tmark & \cmark & \cmark & \cmark \\
Recency                                 & \xmark & \xmark & \xmark & \xmark & \xmark & \xmark & \xmark & \xmark \\
Homophily                               & \tmark & \tmark & \cmark & \tmark & \xmark & \tmark & \tmark & \xmark \\
Preferential Attachment & \cmark & \cmark & \cmark & \cmark & \cmark & \cmark & \cmark & \cmark \\
\bottomrule
\end{tabular}

\medskip
\caption{\capheader{Summary of Results.}
We show the results of our experiments, which evaluate whether eight state-of-the-art graph learning models (columns) learn important characteristics of temporal graphs (rows). 
A \cmark{} indicates success, \xmark{} failure, and \tmark{} limited ability to learn a characteristic.
Overall, we find consistent limitations in these models, such as their inability to distinguish directions of edges or lack of emphasis on recently active edges when predicting future links. At the same time, we find that models consistently learn preferential attachment to popular nodes in link formation and identify differences in their capacity to learn other characteristics.
}
\label{tab:results_summary}
\end{table}

	\subheader{Broader Implications.}
	The results highlight that conventional benchmark evaluations provide only partial insight into model behavior. Aggregate performance metrics do not reveal which underlying graph characteristics drive model predictions, and our analysis shows that models can rely on substantially different signals, even within the same prediction task. By isolating individual graph characteristics, our framework complements performance-oriented benchmarks with an interpretability-driven perspective on model evaluation.
	
	This is particularly important in light of recent work showing that simple heuristics based on recency and popularity can perform competitively with state-of-the-art TGL models \citep{cornell_power_2025}. Our results suggest that such benchmark performance can conflate different predictive signals: while all evaluated models capture popularity-related preferential attachment, none consistently captures recency. Evaluations that isolate individual graph characteristics can therefore help identify which signals models actually learn.
	
	From a practical perspective, our results can support a more informed choice of TGL models. Rather than selecting models solely based on aggregate benchmark performance, practitioners can consider whether a model captures the characteristics relevant to their application. 
	Importantly, we do not argue that all characteristics should be learned in all settings. Depending on the task, some may be irrelevant or even undesirable.
	However, awareness of the learning capabilities of TGL models is crucial for selecting and deploying them in a reliable and context-appropriate manner.
	
	
	
	\subheader{Limitations.}
	Our study has several limitations. First, we conducted an extensive set of experiments involving thousands of model runs across eight graph characteristics. Given the computational demands of these experiments, the number of datasets and graph configurations is necessarily limited, and we do not argue that the selected datasets are representative of all temporal graphs.
	
	Second, our experiments focus on isolated graph characteristics, whereas real-world temporal graphs often exhibit multiple interacting characteristics. This design choice was deliberate: by isolating individual characteristics, we aimed to reduce confounding effects and make it clearer which signals models use for prediction. However, interactions between characteristics may affect model behavior in practice and remain an important direction for future work.
	
	Third, our analysis is based on the \texttt{DyGLib} \citep{yu_better_2024} implementations of the evaluated models. This ensures a consistent experimental framework, but also means that our categorization and conclusions reflect these implementations rather than all details of the original model descriptions. This is particularly relevant for interpreting connections between model design and learned characteristics. In addition, although the evaluated models cover several widely used TGL paradigms, they do not exhaust the space of possible temporal graph architectures.

	Finally, we relied on hyperparameter settings that proved effective on the empirical datasets used to probe density, direction, and temporal granularity. Especially for the synthetic experiments used to test other graph characteristics, we cannot rule out that different hyperparameter choices might influence results. 
	However, additional ablation studies with varied hyperparameters in \Cref{ap:hyperparameters} produced no substantial differences.

	\section{Conclusion}

	In this work, we introduced a framework for evaluating whether temporal graph learning models capture eight intuitive graph characteristics. Applying this framework to eight widely used TGL models, we showed that models differ substantially in the characteristics they capture, revealing behaviors that are not visible from aggregate benchmark performance alone.
	
	Future work could extend the framework to additional graph characteristics, combinations of characteristics, and temporal graph models outside the continuous-time setting considered here. Moreover, the limitations identified in this study provide concrete starting points for developing models that capture relevant graph characteristics more reliably.
	

	\FloatBarrier
	
	\printbibliography
	
	\newpage
	\appendix
	
	\FloatBarrier
	\section{Additional Details on Experimental Setup}
	
	\subsection{Details of Empirical Datasets}\label{ap:datasets}
	In our experiments with empirical data, we consider the following datasets:
	\begin{enumerate}
		\item \emph{Enron} \citep{shetty_enron_2004} is a network modeling email communication within the Enron corporation between 2000 and 2002. Nodes correspond to individuals that worked at Enron, and an edge $(u,v,t)$ indicates that an email was sent from individual $u$ to individual $v$ at time $t$.
		\item \emph{Bitcoin-Alpha} \citep{kumar_edge_2016} is a whom-trust-whom network of bitcoin users that traded on the platform \url{http://www.btc-aplha.com}. An edge $(u,v,t)$ indicates that user $u$ assigned a rating to user $v$ at time $t$. Each edge further contains the rating score which ranges from $-10$ to $10$. 
		\item \emph{UCI} \citep{panzarasa_patterns_2009} represents an online community of students at the University of California, Irvine. Within this community, students could search each other's profiles and message each other. The network is then modeled based on messages sent in the period from April to October 2004. An edge $(u,v,t)$ indicates that user $u$ sent a message to user $v$ at time $t$. 
		\item \emph{Wikipedia} \citep{kumar_predicting_2019} is a heterogeneous network that models one month of edits on Wikipedia pages. An edge $(u,v, t)$ indicates that editor $u$ edited page $v$ at time $t$.
	\end{enumerate}
	
\begin{figure}[b!]
		\includegraphics[width=\textwidth]{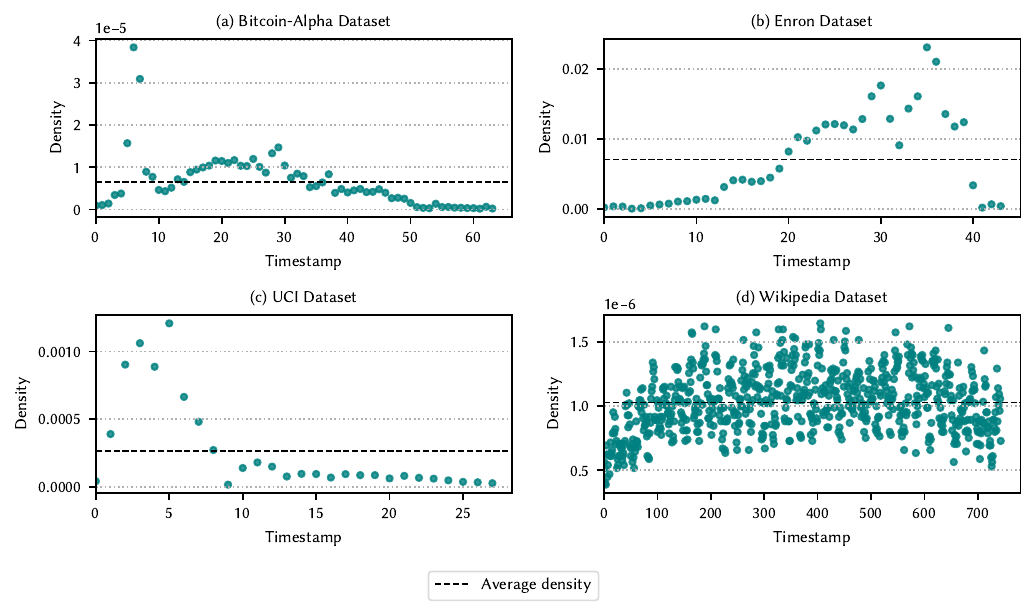}
		\caption{\capheader{Densities of Empirical Datasets.}
			For the discrete versions of the empirical datasets under study, we show the densities across time steps.
		} 
		\label{fig:dataset_densities}
	\end{figure}

	We depict how the densities of these datasets evolve over time in \Cref{fig:dataset_densities}.

	\newpage
	\subsection{Choice of Hyperparameters}\label{ap:hyperparameters}

	
	Hyperparameter choices are included in Table \ref{tab:hps}. For the datasets and models we used, \citet{yu_better_2024} found recent sampling of neighbors to always be the most effective. There was more variation in the other parameters depending on the dataset and model combination. The values fixed here give very similar average precision scores when following the DyGLib evaluation procedure compared to the optimal dataset specific hyperparameters specified in the paper (see Table \ref{tab:granularity_ap}).
	
	\begin{table}[h]

\caption{\capheader{Hyperparameter Settings.} We generally use fixed hyperparameter settings across all experiments, for each model.
}
\label{tab:hps}
\medskip
\small
\centering
\begin{tabular}{ll}
\toprule
\textbf{Model} & \textbf{Hyperparameters} \\
\midrule
\textit{All models} & Learning rate: 1e-4 \\
 & Batch size: 200 \\
 & Loss: BCELoss \\
 & Optimizer: Adam \\
 & Dropout: 0.0 (except for \tgn{} with persistence: 0.2) \\
 & Max epochs: 300 (100 for preferential attachment) \\
 & Early stopping tolerance: 1e-6 \\
 & Early stopping patience: 20 \\
 & Time feature dimension: 100 \\
 & Number of neighbors: 20 \\
 & Time gap for neighbors: 2000 \\
 & Dimension of output representation: 172 \\
 \midrule
 \textbf{CAWN} & Number of walk heads: 8 \\
  & Position feature dimension: 172\\
  & Walk length: 2\\
  \midrule
 \textbf{DyGFormer} & Number of layers: 2\\
  & Number of heads: 2\\
  & Channel embedding dimension: 50\\
  & Patch size: 2\\
  & Max input sequence length: 64\\
  \midrule
 \textbf{DyRep}, \textbf{JODIE} and \textbf{TGN} & Number of layers: 1\\
 & Number of heads: 2\\
 & Dimension of node memory: 172\\
 \midrule
 \textbf{GraphMixer} & Number of layers: 2\\
 & Number of heads: 2\\
 & Number of tokens: 20\\
 \midrule
 \textbf{TCL} & Number of layers: 2\\
 & Number of heads: 2\\
 & Number of depths: 21\\
 \midrule
 \textbf{TGAT} & Number of layers: 2\\
 & Number of heads: 2\\
\bottomrule
\end{tabular}
\end{table}

	\FloatBarrier

	\newpage
	\section{Additional Plots and Tables}\label{ap:results}
	
	In the following, we present and discuss additional results from our experiments regarding \emph{temporal granularity}, \emph{direction}, \emph{persistence}, \emph{periodicity}, \emph{homophily}, and \emph{preferential attachment}.

	\subsubsection*{\circled{1} Temporal Granularity}
	\Cref{tab:granularity_ap} complements the results from the main text in \Cref{tab:time_granularity} by providing average precision scores. Trends are overall in line with those from the main text, and further, the average precision scores achieved on continuous datasets are comparable to those reported by \citet{yu_better_2024}.
	
	In \Cref{tab:granularity_duplicate_results}, we further analyze the impact of removing duplicate edges that result from coarsening time steps on the performance in terms of ROC-AUC score.
	Here we see that the memory-based \dyrep{}, \jodie{}, and \tgn{} models generally deteriorate in performance when duplicate edges are removed while for all other models, this impact is much more dataset-dependent.

	\begin{table*}[h]
\centering
\resizebox{\linewidth}{!}{
\begin{tabular}{lcccccccccccc}
\toprule
 & \multicolumn{3}{c}{Bitcoin-Alpha}
 & \multicolumn{3}{c}{Enron}
 & \multicolumn{3}{c}{UCI}
 & \multicolumn{3}{c}{Wikipedia} \\
\cmidrule(lr){2-4} \cmidrule(lr){5-7}
\cmidrule(lr){8-10} \cmidrule(lr){11-13}
 & Cont. & Disc. & Flat
 & Cont. & Disc. & Flat
 & Cont. & Disc. & Flat
 & Cont. & Disc. & Flat \\
\midrule
\textbf{CAWN      } &   0.745 &   0.986 & 0.675 & 0.958 &   0.945 & 0.497 & 0.976 &   0.952 & 0.694 &     0.992 &   0.986 & 0.589 \\
\textbf{DyGFormer } &   0.862 &   0.990 & 0.759 & 0.951 &   0.946 & 0.753 & 0.974 &   0.959 & 0.576 &     0.991 &   0.983 & 0.531 \\
\textbf{DyRep     } &   0.896 &   0.866 & 0.927 & 0.872 &   0.654 & 0.556 & 0.921 &   0.875 & 0.817 &     0.968 &   0.944 & 0.822 \\
\textbf{GraphMixer} &   0.997 &   0.994 & 0.597 & 0.946 &   0.860 & 0.507 & 0.986 &   0.918 & 0.545 &     0.979 &   0.950 & 0.494 \\
\textbf{JODIE     } &   0.999 &   0.999 & 0.951 & 0.920 &   0.815 & 0.642 & 0.955 &   0.939 & 0.828 &     0.972 &   0.958 & 0.817 \\
\textbf{TCL       } &   0.896 &   0.991 & 0.693 & 0.861 &   0.860 & 0.577 & 0.966 &   0.946 & 0.649 &     0.979 &   0.957 & 0.578 \\
\textbf{TGAT      } &   0.810 &   1.000 & 0.657 & 0.770 &   0.860 & 0.524 & 0.888 &   0.929 & 0.550 &     0.972 &   0.963 & 0.506 \\
\textbf{TGN       } &   0.995 &   0.999 & 0.954 & 0.894 &   0.888 & 0.551 & 0.984 &   0.944 & 0.820 &     0.987 &   0.974 & 0.845 \\
\bottomrule
\end{tabular}
}
\caption{\capheader{Temporal Granularity:} \textit{impact of granularity of timestamps on performance.} We show the mean average precision of temporal graph learning models on benchmark test sets for varying time granularity. These reflect the same patterns seen for ROC-AUC in Table \ref{tab:time_granularity}, with performance on the continuous datasets comparable to the values found by \citet{yu_better_2024}.}
\label{tab:granularity_ap}
\end{table*}

\begingroup
\setlength{\tabcolsep}{2.8pt}
\begin{table}[h]
\centering
\begin{tabular}{@{}l llll G llll G llll@{}}
\toprule
&
\multicolumn{4}{c}{Enron}
&
&
\multicolumn{4}{c}{UCI}
&
&
\multicolumn{4}{c}{Wikipedia}
\\
\cmidrule(lr){2-5}
\cmidrule(lr){7-10}
\cmidrule(lr){12-15}
&
\multicolumn{2}{c}{Discrete}
&
\multicolumn{2}{c}{Flattened}
&
&
\multicolumn{2}{c}{Discrete}
&
\multicolumn{2}{c}{Flattened}
&
&
\multicolumn{2}{c}{Discrete}
&
\multicolumn{2}{c}{Flattened}
\\
\cmidrule(lr){2-3}
\cmidrule(lr){4-5}
\cmidrule(lr){7-8}
\cmidrule(lr){9-10}
\cmidrule(lr){12-13}
\cmidrule(lr){14-15}
&
Orig. & Dedup.
&
Orig. & Dedup.
&
&
Orig. & Dedup.
&
Orig. & Dedup.
&
&
Orig. & Dedup.
&
Orig. & Dedup.
\\
\midrule
\textbf{CAWN} & 0.903 & 0.941\sig & 0.630 & 0.518\sig &  & 0.971 & 0.953\sig & 0.671 & 0.684 &  & 0.988 & 0.983\sig & 0.625 & 0.571 \\
\textbf{DyGFormer} & 0.922 & 0.939\sig & 0.639 & 0.751\sig &  & 0.972 & 0.960\sig & 0.463 & 0.532 &  & 0.986 & 0.980\sig & 0.507 & 0.495 \\
\textbf{DyRep} & 0.830 & 0.732\sig & 0.810 & 0.576\sig &  & 0.935 & 0.889\sig & 0.916 & 0.834\sig &  & 0.965 & 0.943\sig & 0.958 & 0.829\sig \\
\textbf{GraphMixer} & 0.823 & 0.871\sig & 0.459 & 0.451 &  & 0.944 & 0.917\sig & 0.518 & 0.477 &  & 0.955 & 0.945\sig & 0.619 & 0.435\sig \\
\textbf{JODIE} & 0.911 & 0.860\sig & 0.853 & 0.701\sig &  & 0.964 & 0.946\sig & 0.941 & 0.845\sig &  & 0.971 & 0.955\sig & 0.954 & 0.829\sig \\
\textbf{TCL} & 0.832 & 0.870\sig & 0.554 & 0.580 &  & 0.969 & 0.949\sig & 0.677 & 0.624 &  & 0.960 & 0.952\sig & 0.625 & 0.549 \\
\textbf{TGAT} & 0.844 & 0.875\sig & 0.488 & 0.496 &  & 0.925 & 0.926 & 0.491 & 0.463 &  & 0.966 & 0.960\sig & 0.522 & 0.506 \\
\textbf{TGN} & 0.912 & 0.889\sig & 0.904 & 0.598\sig &  & 0.971 & 0.948\sig & 0.915 & 0.794\sig &  & 0.979 & 0.971\sig & 0.967 & 0.848\sig \\
\bottomrule
\end{tabular}
\medskip
\caption{\capheader{Impact of Removing Duplicate Edges.} 
We show mean ROC-AUC scores of the evaluated graph learning models on the given datasets across the discrete and flat settings.
For each setting, we compare performance using the original edge set (\emph{Orig.}) with a deduplicated variant (\emph{Dedup.}), in which duplicate edges introduced by timestamp coarsening are removed from both the training and test sets.
A $\sig$ indicates that the given AUC score differs from the score without deduplication to a degree that is statistically significant with respect to a permutation test.
\dyrep{}, \jodie{}, and \tgn{} consistently deteriorate in performance when duplicate edges are removed; for all other models, the impact of edge removal on performance is largely dataset-dependent.}
\label{tab:granularity_duplicate_results}
\end{table}

\endgroup
	
	\FloatBarrier
	\newpage

	\subsubsection*{\circled{2} Direction}
\begin{figure}[h]
		\includegraphics[width=\textwidth]{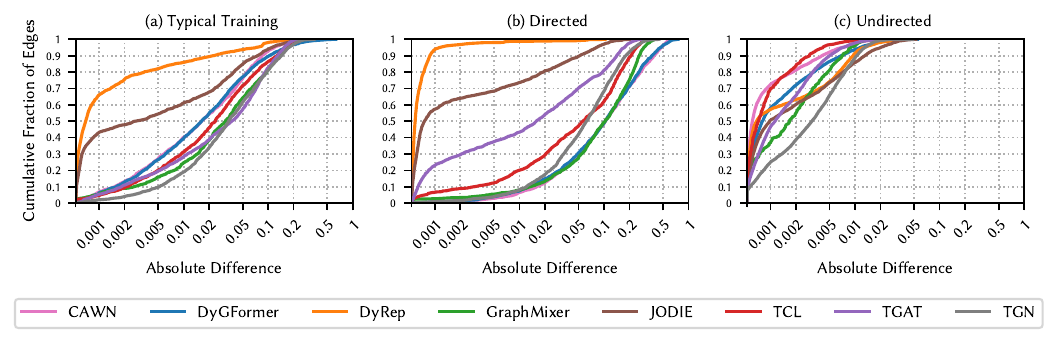}
		\caption{\capheader{Direction:} \emph{ability of graph learning models to distinguish edge directions.} Additional results for the Enron dataset. \dyrep{} and \jodie{} particularly struggle with distinguishing the two edge directions.} 
		\label{fig:direction_enron}
	\end{figure}

\begin{figure}[h]
		\includegraphics[width=\textwidth]{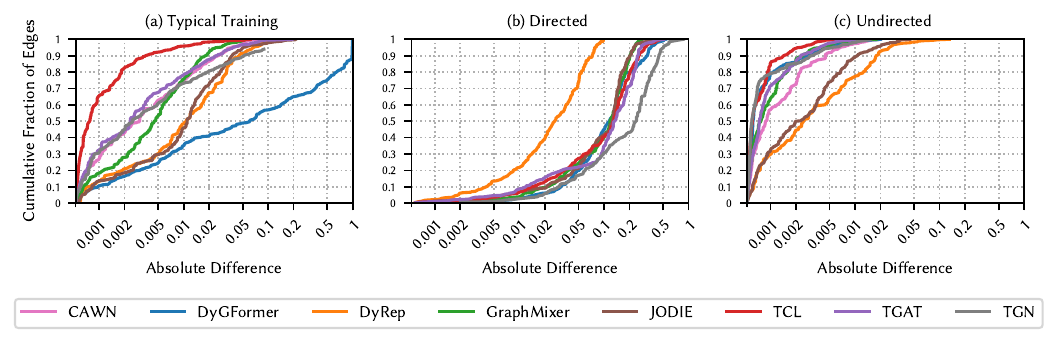}
		\caption{\capheader{Direction:} \emph{ability of graph learning models to distinguish edge directions.} Additional results for the Bitcoin-Alpha dataset. Compared to the results on the UCI dataset in \Cref{fig:direction_uci}, some models, apart from \dygformer{}, are less able to distingush edge pairs in the original dataset, but with improved results when reverse edges are explicitly provided as negatives. This might reflect the fact that this dataset never contains an edge in both directions at the same time step.} 
		\label{fig:direction_bca}
	\end{figure}

	Figures \ref{fig:direction_enron} and \ref{fig:direction_bca} show results for the Enron and Bitcoin-Alpha datasets.
	In general, results are in line with those observed on the UCI dataset, cf. \Cref{fig:direction_uci}: for the original training edges, most edge probabilities are very symmetric, but there are also higher values, specifically for \dygformer{}.
	In the \emph{directed} setting, providing a negative reverse training edge for each positive training edge tends to increase differences between positive test edges and their negative reverse, while always providing both directions in the \emph{undirected} setting generally yields much more symmetric predictions.
	Yet, on Enron, it appears that in the directed setting, these differences tend to decrease for \dyrep{}, \jodie{}, and \tgat{}.
	This is, however, likely due to the overall higher ratio of negative to positive sample pushing many edge probabilities toward 0. As observed in our experiments on \emph{density}, an increase in negative samples will generally decrease the probabilities of edges being predicted as positive.
	We can also see in Table \ref{tab:direction} that the predicted edge probabilities in the \emph{directed} setting are generally lower.

	\newpage
	\begin{table*}[ht]
\resizebox{\textwidth}{!}{
\begin{tabular}{llcccccccccccc}
\toprule
 &  & \multicolumn{4}{c}{Bitcoin-Alpha} & \multicolumn{4}{c}{Enron} & \multicolumn{4}{c}{UCI} \\
 \cmidrule(lr){3-6} \cmidrule(lr){7-10} \cmidrule(lr){11-14}
 \makecell{Dataset\\ variant} & Model
 & Edges & \makecell{Both\\ directions} & \makecell{Reverse\\ edges} & Other
 & Edges & \makecell{Both\\ directions} & \makecell{Reverse\\ edges} & Other
 & Edges & \makecell{Both\\ directions} & \makecell{Reverse\\ edges} & Other \\
 \midrule
\multirow{8}{*}{Original}   & CAWN          & 0.92 & - & 0.92 & 0.03 & 0.80 & 0.89 & 0.79 & 0.23 & 0.73 & 0.83 & 0.72 & 0.10 \\
                            & DyGFormer     & 0.87 & - & 0.67 & 0.03 & 0.79 & 0.89 & 0.78 & 0.19 & 0.77 & 0.86 & 0.80 & 0.13 \\
                            & DyRep         & 0.83 & - & 0.83 & 0.16 & 0.62 & 0.63 & 0.62 & 0.54 & 0.73 & 0.80 & 0.74 & 0.31 \\
                            & GraphMixer    & 0.91 & - & 0.91 & 0.03 & 0.71 & 0.77 & 0.69 & 0.42 & 0.71 & 0.81 & 0.71 & 0.15 \\
                            & JODIE         & 0.88 & - & 0.88 & 0.01 & 0.60 & 0.63 & 0.60 & 0.33 & 0.71 & 0.81 & 0.70 & 0.08 \\
                            & TCL           & 0.93 & - & 0.93 & 0.03 & 0.68 & 0.74 & 0.67 & 0.40 & 0.70 & 0.81 & 0.70 & 0.12 \\
                            & TGAT          & 0.94 & - & 0.93 & 0.02 & 0.56 & 0.63 & 0.55 & 0.34 & 0.82 & 0.86 & 0.81 & 0.21 \\
                            & TGN           & 0.92 & - & 0.92 & 0.02 & 0.72 & 0.80 & 0.71 & 0.37 & 0.83 & 0.88 & 0.81 & 0.10 \\
\midrule                                                                
\multirow{8}{*}{Undirected} & CAWN          & 0.94 & - & 0.94 & 0.02 & 0.87 & 0.95 & 0.87 & 0.33 & 0.77 & 0.84 & 0.77 & 0.13 \\
                            & DyGFormer     & 0.94 & - & 0.94 & 0.02 & 0.83 & 0.92 & 0.83 & 0.22 & 0.80 & 0.85 & 0.82 & 0.13 \\
                            & DyRep         & 0.84 & - & 0.84 & 0.15 & 0.65 & 0.67 & 0.65 & 0.53 & 0.79 & 0.85 & 0.79 & 0.26 \\
                            & GraphMixer    & 0.93 & - & 0.93 & 0.01 & 0.65 & 0.70 & 0.65 & 0.41 & 0.80 & 0.84 & 0.80 & 0.17 \\
                            & JODIE         & 0.90 & - & 0.90 & 0.01 & 0.57 & 0.59 & 0.57 & 0.33 & 0.71 & 0.79 & 0.71 & 0.10 \\
                            & TCL           & 0.94 & - & 0.94 & 0.02 & 0.57 & 0.61 & 0.57 & 0.37 & 0.77 & 0.85 & 0.77 & 0.13 \\
                            & TGAT          & 0.93 & - & 0.93 & 0.02 & 0.58 & 0.64 & 0.58 & 0.34 & 0.79 & 0.83 & 0.79 & 0.20 \\
                            & TGN           & 0.94 & - & 0.94 & 0.02 & 0.67 & 0.75 & 0.67 & 0.37 & 0.82 & 0.87 & 0.82 & 0.14 \\
\midrule                                                                
\multirow{8}{*}{Directed}   & CAWN          & - & - & - & - & 0.56 & 0.66 & 0.52 & 0.15 & 0.50 & 0.60 & 0.50 & 0.09 \\
                            & DyGFormer     & 0.70 & - & 0.61 & 0.08 & 0.64 & 0.71 & 0.59 & 0.15 & 0.62 & 0.70 & 0.63 & 0.13 \\
                            & DyRep         & 0.47 & - & 0.45 & 0.13 & 0.34 & 0.34 & 0.34 & 0.34 & 0.53 & 0.59 & 0.53 & 0.19 \\
                            & GraphMixer    & 0.68 & - & 0.61 & 0.03 & 0.50 & 0.54 & 0.48 & 0.30 & 0.52 & 0.59 & 0.50 & 0.12 \\
                            & JODIE         & 0.66 & - & 0.58 & 0.01 & 0.36 & 0.37 & 0.36 & 0.24 & 0.52 & 0.59 & 0.50 & 0.05 \\
                            & TCL           & 0.71 & - & 0.62 & 0.02 & 0.46 & 0.49 & 0.44 & 0.28 & 0.52 & 0.61 & 0.51 & 0.11 \\
                            & TGAT          & 0.70 & - & 0.59 & 0.02 & 0.39 & 0.42 & 0.38 & 0.28 & 0.56 & 0.63 & 0.56 & 0.13 \\
                            & TGN           & 0.68 & - & 0.52 & 0.02 & 0.49 & 0.54 & 0.46 & 0.25 & 0.64 & 0.68 & 0.60 & 0.07 \\
\bottomrule
\end{tabular}
}
\caption{\capheader{Direction:} \emph{ability of graph learning models to distinguish directions of edges.}
The average probability given to each edge by group, for each model and dataset variant combination for the Enron, Bitcoin-Alpha and UCI datasets, with 1:1 negative edges in training. There are no edges in Bitcoin-Alpha which appear in both directions in the testing timestep. There is generally very little difference in the average probability given to true edges and their reverse edges which should not appear. However, edges which should appear in both directions generally receive a higher average probability than those which should appear in only a single direction.}\label{tab:direction}
\end{table*}

	Table \ref{tab:direction} further illustrates additional limitations in the ability of models to learn direction of edges.
	Generally, it would be desirable if the non-appearing reverses of positive edges would differ more strongly in their predicted probabilities from the true positive edges.
	When comparing the average probability scores of reverse positive edges in the \emph{original} setting, which are mostly between 0.7-0.8, we find a stark difference to those of all other negative edges  which tend to vary between 0.1-0.2.
	Even more strikingly, for some models, the reverses of positive edges receive higher average probability scores than their true positive counterparts. 
	While it is plausible that the probability of a signal between nodes which had a previous signal in either direction comes out higher than the probability of a signal between nodes which were never in contact, the reverse edge probabilities should still be more distinguishable from the true edges, and scores between 0.4-0.5 would appear to be more sensible estimations.
	Training in the \emph{directed} setting appears to alleviate this to some degree specifically on Bitcoin-Alpha.
	For the other datasets, the averages appear less affected, however, this may also be due to many positive test edges having been observed in both directions during training, given the reciprocal nature of these communication networks, and the number of temporal links of these datasets in relation to the number of nodes.
	
	\FloatBarrier
	
	\newpage
	\subsubsection*{\circled{4} Persistence}
\begin{figure}[h]
		\includegraphics[width=\textwidth]{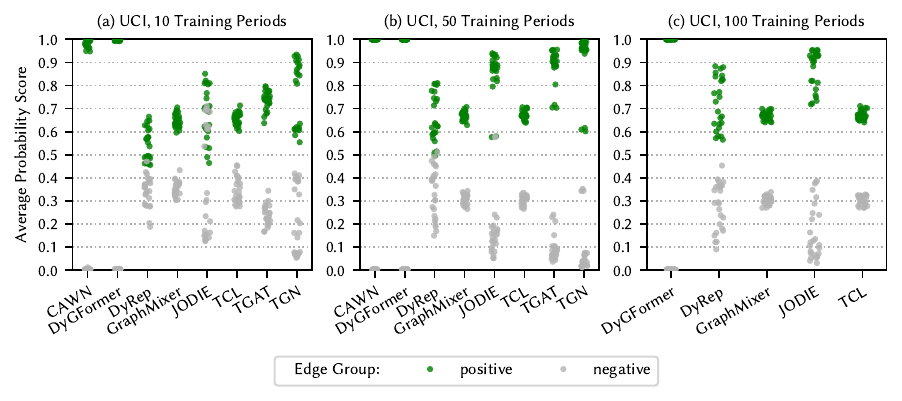}
		\caption{\capheader{Persistence:} \emph{ability of models to learn persistent snapshots.} Additional results using snapshots from the UCI dataset after training for $T\in\{10,50,100\}$ time steps. Only \cawn{} and \dygformer{} appear to reproduce the fixed graph pattern early on with high confidence, while all other models generally discriminate positive and negative edges after a sufficient amount of time steps. For $T=100$ we only show partial results, models that are not shown already separate after training for $T=50$ time steps.
		}
		\label{fig:persistence_uci}
	\end{figure}

\begin{figure}[h]
		\includegraphics[width=\textwidth]{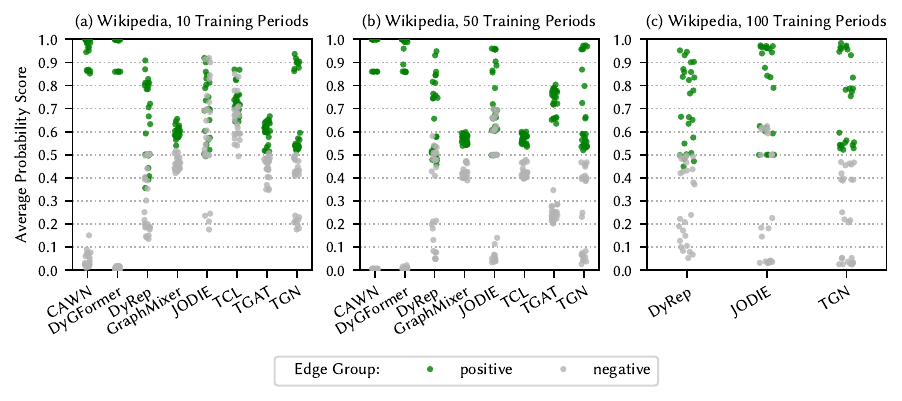}
		\caption{\capheader{Persistence:} \emph{ability of models to learn persistent snapshots.} Additional results using snapshots from the Wikipedia dataset after training for $T\in\{10,50,100\}$ time steps. Again, only \cawn{} and \dygformer{} appear to reproduce the fixed graph pattern early on with high confidence, while most other models generally discriminate positive and negative edges after a sufficient amount of time steps. For $T=100$ we only show partial results, models that are not shown already separate after training for $T=50$ time steps. At this stage, \dyrep{} and \jodie{} still don't fully separate positive and negative edges, although trends tend to improve compared to $T=50$.
		}
		\label{fig:persistence_wiki}
	\end{figure}

\begin{figure}[h]
		\includegraphics[width=\textwidth]{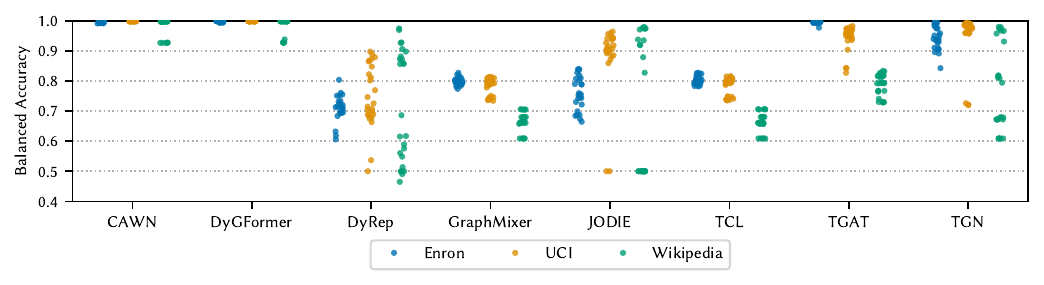}
		\caption{\capheader{Persistence:} \emph{ability of models to learn persistent snapshots.} We show balanced accuracies of all models across all datasets and training seeds, with datasets having seen the persistent graph for 50 time steps. Only \cawn{} and \dygformer{} appear to consistently learn persistent networks, other models struggle to varying degrees.
		}
		\label{fig:persistence_accuracies}
	\end{figure}

	\Cref{fig:persistence_uci} and \Cref{fig:persistence_wiki} show extra results from training on persistent snapshots of the UCI and Wikipedia datasets. These results are overall in line with the results on the data created from the Enron dataset, as reported in \Cref{fig:persistence_enron} in the main text—--only that on Wikipedia, models generally require more time to separate positive and negative edges, with \dyrep{} and \jodie{} still not achieving full separation after $T=100$ time steps.
	\Cref{fig:persistence_accuracies} shows the balanced accuracies across all datasets, which, for all models except \cawn{} and \dygformer{}, are surprisingly low considering the trivial nature of the given prediction task.

	\clearpage

	\FloatBarrier
	\subsubsection*{\circled{5} Periodicity}
	
\begin{figure}[h]
		\centering
		\includegraphics{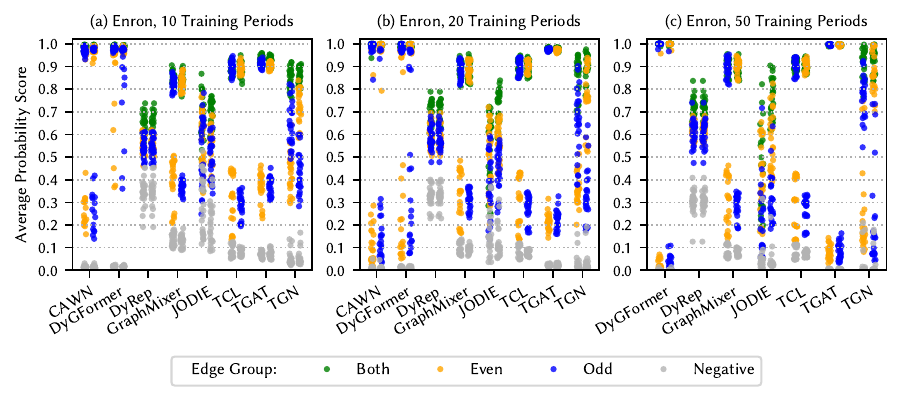}
		\caption{\capheader{Periodicity:} \emph{ability of models to learn periodically
				repeated edges.} Additional results using snapshots from the Enron dataset after training for 10, 20 and 50 periods. Except for \dyrep{} and \jodie{}, all models appear to eventually reproduce the periodic pattern.}
		\label{fig:periodicity_enron}
	\end{figure}

\begin{figure}[h]
		\centering
		\includegraphics{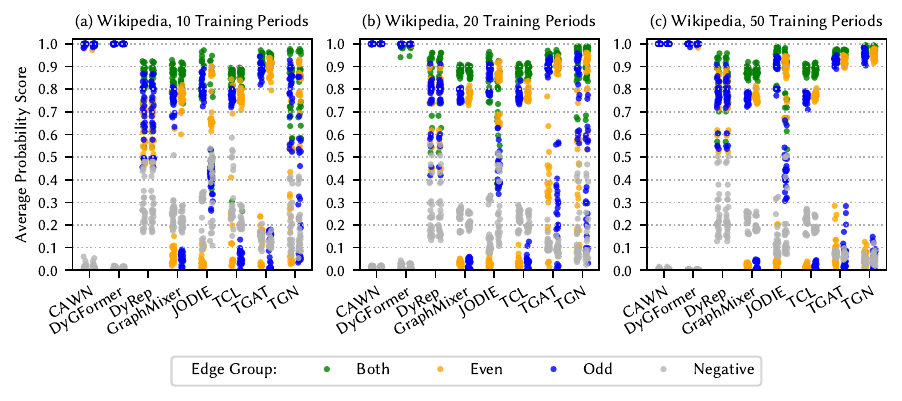}
		\caption{\capheader{Periodicity:} \emph{ability of models to learn periodically
				repeated edges.} Additional results using snapshots from the Wikipedia dataset after learning for 10, 20, and 50 periods. Except for \cawn{}, \dygformer{} and \dyrep{}, all models appear to eventually learn the periodic pattern.}
		\label{fig:periodicity_wiki}
\end{figure}

	Figures \ref{fig:periodicity_enron} and \ref{fig:periodicity_wiki} show additional results with snapshots from the Enron and Wikipedia datasets. 
	Results are mostly in line with those observed on the UCI dataset (see \Cref{fig:periodicity_uci}).
	Yet, we see a few notable differences on Enron for \cawn{}, \dygformer{}, and \jodie{}.
	First, \cawn{} and \dygformer{} appear to distinguish different snapshots to some degree, which is not the case for the other datasets.
	This may indicate that under some circumstances, these methods can learn periodic patterns.
	Conversely, \jodie{} struggles at learning the periodic patterns on this specific dataset, and also \tgn{} takes longer to pick up the periodic pattern. 
	
	On Wikipedia, \cawn{} and \dygformer{} again show the behavior from UCI, not distinguishing edges seen at odd and even time steps.
	Except \dyrep{}, all other models pick up on the periodic pattern, with \jodie{} being the slowest at doing so.
	
\begin{figure}[h]
\centering
\includegraphics{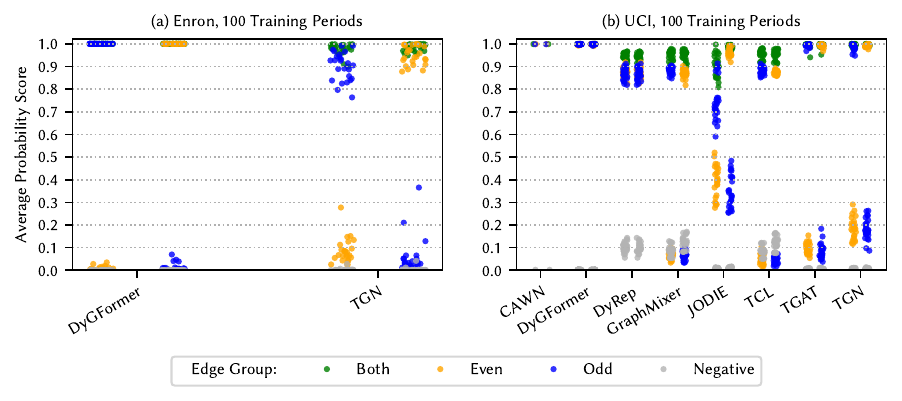}
\caption{\capheader{Periodicity:} \emph{ability of models to learn periodically
		repeated edges.} Additional results using snapshots from the Enron and UCI Wikipedia dataset after learning for 100 periods. 
	We observe that on Enron, \dygformer{} and \tgn{} eventually distinguish odd and even time steps in a clear manner. On UCI, the trends from training for 50 periods are also reinforced.}
\label{fig:periodicity_100}
\end{figure}

	We further display results for training selected models on Enron and UCI for 100 periods in \Cref{fig:periodicity_100}. We observe that specifically on Enron, \dygformer{} and \tgn{}, which have not achieved clear separation of odd and even time steps after 50 periods, have now achieved this separation. Similarly, on UCI, the trends observed after 50 training periods are confirmed.
	
\begin{figure}[h]
\centering
\includegraphics{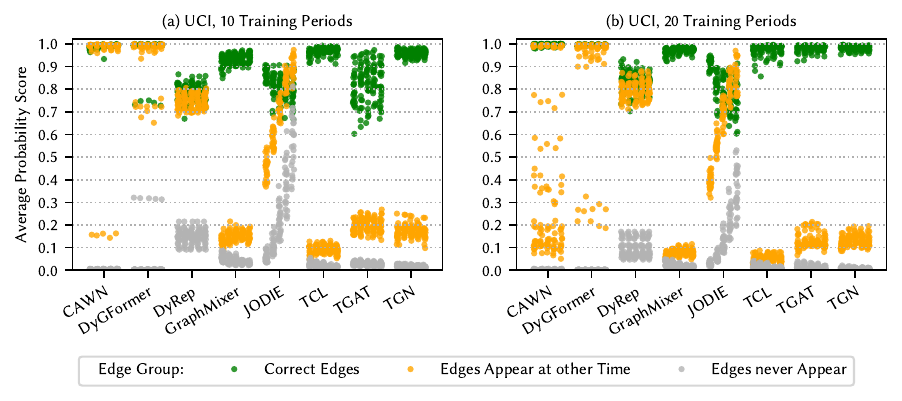}
\caption{\capheader{Periodicity:} \emph{ability of models to learn periodically
		repeated edges.} Additional results using snapshots from Enron with period 5, which are repeated for 10 and 20 training periods. For each of the five time steps, edges are grouped by whether they are (i) positive edges at this time step, (ii) negative edges here, but positive at one or more other time steps, or (iii) negative at all time steps. Except for \cawn{}, \dygformer{}, \dyrep{}, and \jodie{}, all models appear to reproduce the periodic pattern eventually.}
\label{fig:periodicity_uci5}
\end{figure}

\begin{figure}[h]
\centering
\includegraphics{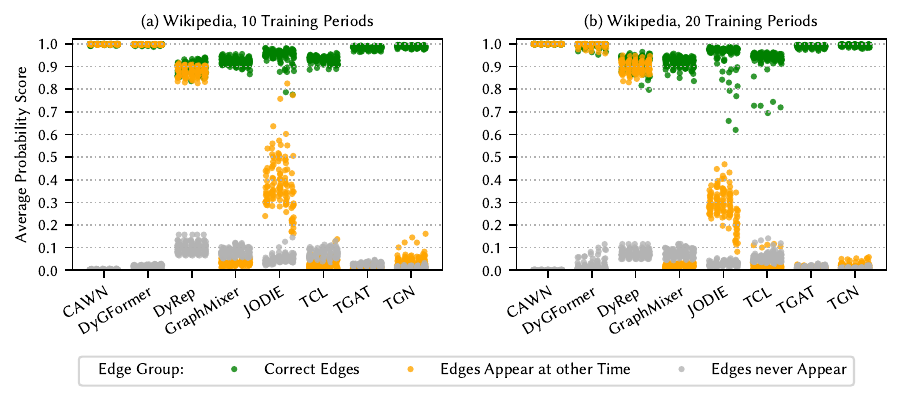}
\caption{\capheader{Periodicity:} \emph{ability of models to learn periodically
		repeated edges.} Additional results using snapshots from Wikipedia with period 5, which are repeated for 10 and 20 training periods. For each of the five time steps, edges are grouped by whether they are (i) positive edges at this time step, (ii) negative edges here, but positive at one or more other time steps, or (iii) negative at all time steps. Except for \cawn{}, \dygformer{}, and  \dyrep{}, all models appear to reproduce the periodic pattern eventually.}
\label{fig:periodicity_wiki5}
\end{figure}

	These general trends are also consistent with the results for period length 5, as depicted in Figures \ref{fig:periodicity_uci5} and \ref{fig:periodicity_wiki5}.
	Regarding these, on UCI, \cawn{} and \dygformer{} show indications of distinguishing edges from different timesteps with more training, while for \jodie{}, there appears to be some skew in the time steps that are learned.
	On Wikipedia, \cawn{}, \dygformer{} and \dyrep{} do not appear to distinguish edges from different time steps, while all other models do so.

\begin{figure}[h]
\includegraphics[width=\textwidth]{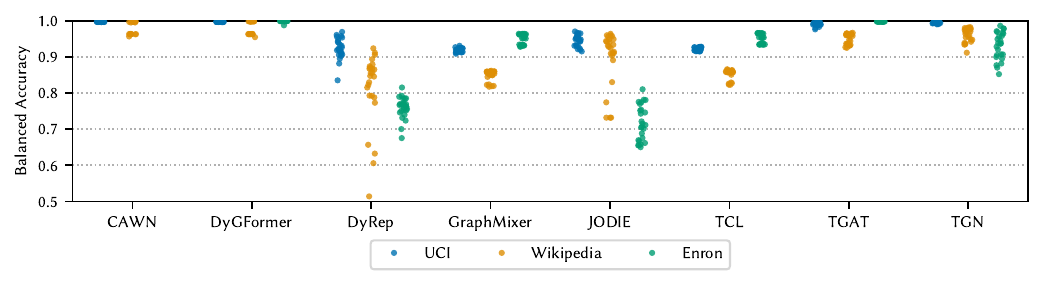}
\caption{\capheader{Periodicity:} \emph{ability of models to learn periodically changing edges.} We show the balanced accuracies of all models across all datasets and training seeds, with averages across odd and even test time steps. All models except for \dyrep{} and \jodie{} appear to perform relatively well at this classification task. 
}
\label{fig:periodicity_accuracies}
\end{figure}

	Finally, \Cref{fig:periodicity_accuracies} shows the balanced accuracies for the period 2 predictions across all datasets.
	Here, it appears that all models except for \dyrep{} and \jodie{} perform well overall.
	
\FloatBarrier

\subsubsection*{\circled{7} Homophily}

Table \ref{tab:homophily_ranks} shows the share of the top-$k$ edges that belong to intra-group and inter-group edges.
Overall, these plots also illustrate how most models are able to distinguish between inter-group links and intra-group links, generally completely placing the minority edge group(s) at the bottom of these rankings.
While the group-wise shares of edges among the top-$k$ edges are sometimes unbalanced in the homophilic SBM, these are often only due to slight differences in average edge probabilities, cf. Table \ref{tab:homophily_probs}.

\begin{table*}[h]
\centering
\resizebox{\textwidth}{!}{
\begin{tabular}{lrrrrrrrrrrrrrrrrrr}
\toprule
 & \multicolumn{9}{c}{Homophilic SBM}
 & \multicolumn{9}{c}{Heterophilic SBM} \\
\cmidrule(lr){2-10} \cmidrule(lr){11-19}
 & \multicolumn{3}{c}{\makecell{Share within\\Top 1000 Edges}}
 & \multicolumn{3}{c}{\makecell{Share within\\Top 10,000 Edges}}
 & \multicolumn{3}{c}{\makecell{Share within\\Top 100,000 Edges}}
 & \multicolumn{3}{c}{\makecell{Share within\\Top 1000 Edges}}
 & \multicolumn{3}{c}{\makecell{Share within\\Top 10,000 Edges}}
 & \multicolumn{3}{c}{\makecell{Share within\\Top 100,000 Edges}} \\
\cmidrule(lr){2-4} \cmidrule(lr){5-7} \cmidrule(lr){8-10}
\cmidrule(lr){11-13} \cmidrule(lr){14-16} \cmidrule(lr){17-19}
\textbf{Edge Group}
 & 0--0 & 0--1 & 1--1
 & 0--0 & 0--1 & 1--1
 & 0--0 & 0--1 & 1--1
 & 0--0 & 0--1 & 1--1
 & 0--0 & 0--1 & 1--1
 & 0--0 & 0--1 & 1--1 \\
\midrule
\textbf{CAWN}       &   0.16 & 0.00 & 0.84 &   0.18 & 0.00 & 0.82 &   0.18 & 0.00 & 0.82 & 0.00 & 1.00 & 0.00 & 0.00 & 1.00 & 0.00 & 0.00 & 1.00 & 0.00 \\
\textbf{DyGFormer}  &   0.44 & 0.00 & 0.56 &   0.40 & 0.00 & 0.60 &   0.45 & 0.00 & 0.55 & 0.00 & 1.00 & 0.00 & 0.00 & 1.00 & 0.00 & 0.00 & 1.00 & 0.00 \\
\textbf{DyRep}      &   0.27 & 0.00 & 0.73 &   0.40 & 0.00 & 0.60 &   0.63 & 0.01 & 0.37 & 0.00 & 1.00 & 0.00 & 0.00 & 1.00 & 0.00 & 0.00 & 1.00 & 0.00 \\
\textbf{GraphMixer} &   0.20 & 0.00 & 0.80 &   0.26 & 0.00 & 0.74 &   0.34 & 0.00 & 0.67 & 0.00 & 1.00 & 0.00 & 0.00 & 1.00 & 0.00 & 0.00 & 1.00 & 0.00 \\
\textbf{JODIE}      &   1.00 & 0.00 & 0.00 &   1.00 & 0.00 & 0.00 &   0.96 & 0.00 & 0.04 & 0.03 & 0.97 & 0.00 & 0.01 & 0.99 & 0.00 & 0.10 & 0.90 & 0.00 \\
\textbf{TCL}        &   0.43 & 0.00 & 0.57 &   0.41 & 0.00 & 0.59 &   0.44 & 0.00 & 0.56 & 0.00 & 1.00 & 0.00 & 0.00 & 1.00 & 0.00 & 0.00 & 1.00 & 0.00 \\
\textbf{TGAT}       &   0.20 & 0.00 & 0.80 &   0.16 & 0.00 & 0.84 &   0.18 & 0.00 & 0.82 & 0.00 & 1.00 & 0.00 & 0.00 & 1.00 & 0.00 & 0.00 & 1.00 & 0.00 \\
\textbf{TGN}        &   0.39 & 0.00 & 0.61 &   0.68 & 0.03 & 0.29 &   0.41 & 0.41 & 0.17 & 0.00 & 1.00 & 0.00 & 0.08 & 0.91 & 0.01 & 0.68 & 0.26 & 0.07 \\
\bottomrule
\end{tabular}
}
\caption{\capheader{Homophily:} \emph{ability of models to reproduce homophily in edge formation.} We train graph learning models on stochastic block models with two groups (0 and 1), with intra-group being five times more likely than inter-group links (homophilic) or the reverse (heterophilic). We depict the ratio of the resulting edge groups among the top $k$ most likely edges.
Models generally prefer homophilic edges over heterophilic edges when trained on the heterophilic data, and the reverse when trained on heterophilic data. 
}
\label{tab:homophily_ranks}
\end{table*}

\subsubsection*{\circled{8} Preferential Attachment}

\begin{figure}[h!]
\centering
\includegraphics[]{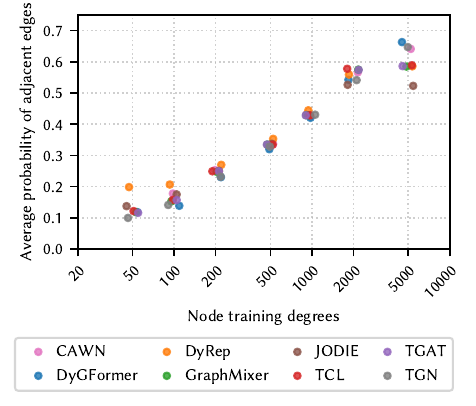}
\caption{\capheader{Preferential Attachment:} \emph{ability of models to reproduce preferential attachment in edge formation.} Additional results on denser BA graphs. While minimum degree is naturally higher, we observe a pattern consistent with the more shallow graphs: edge probabilities increase with degree of its adjacent nodes.
}
\label{fig:pa_dense}
\end{figure}

\Cref{fig:pa_dense} displays results for a denser graph with 4000 edges per time step rather than the 2000 edges used in the main results.
As for the sparser networks presented in \Cref{fig:pa}, we observe that across all models, predicted edge probabilities increase with the degree of the adjacent nodes.

\FloatBarrier
\section{Results with Alternative Hyperparameters}\label{ap:alt_hp}

\citet{yu_better_2024} have already explored the hyperparameter spaces for the models under study with respect to their general performance on a range of empirical datasets \citep{yu_better_2024}. 
Thus, for our hyperparameter ablations, we focus on characteristics that operate on synthetic data.
For persistence and periodicity, we have already conducted hyperparameter ablations on JODIE, TCL and TGN which informed or confirmed our parameter choices.

In the following, we show further ablations of the parameters of these models with respect to the recency and homophily characteristics.
Since extensive experimentation is costly due to the number of different datasets, seeds and models, we restricted the generation and training seeds to three variants rather than five.

Following \citet{yu_better_2024}, for \jodie{} we vary the dropout at intervals of 0.1 from 0.0 to 0.6, resulting in 99 additional models trained. For \tcl{}, we switch to using uniform neighbor sampling and vary dropout at intervals of 0.2 from 0.0 to 0.6 (with fewer variants than \jodie{} due to the larger overall search space).

\begin{figure}[h]
\centering
\includegraphics[width=\textwidth]{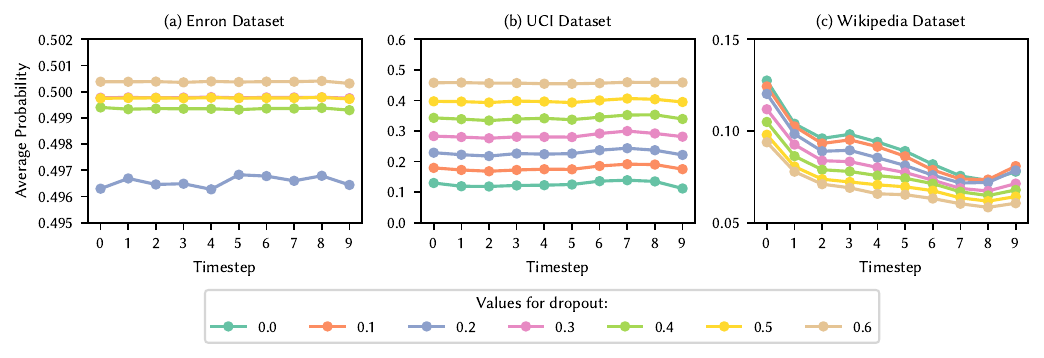}
\caption{\capheader{Recency:} results for different hyperparameters for JODIE. Note the very restricted y-axis range for Wikipedia.}
\label{fig:hp_recency_jodie}
\end{figure}

\begin{figure}[h]
\includegraphics[width=\textwidth]{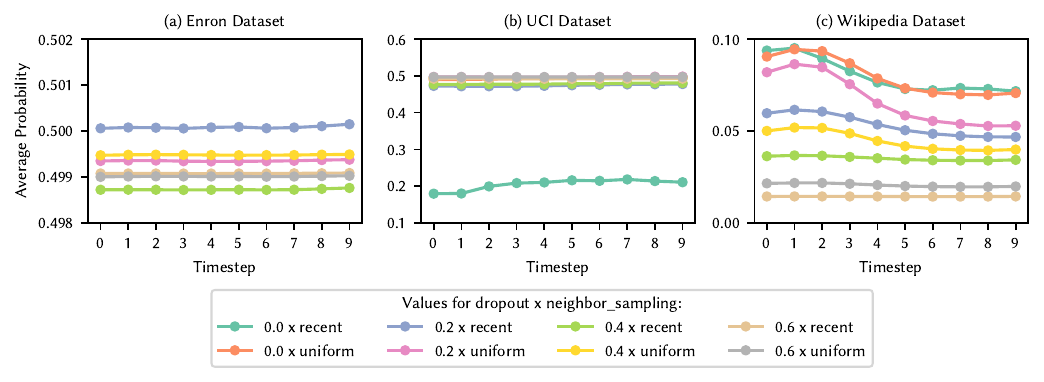}
\centering\caption{\capheader{Recency:} results for different hyperparameters for TCL. Note the very restricted y-axis range for Wikipedia.}
\label{fig:hp_recency_tcl}
\end{figure}

\begin{figure}[h]
\includegraphics[width=\textwidth]{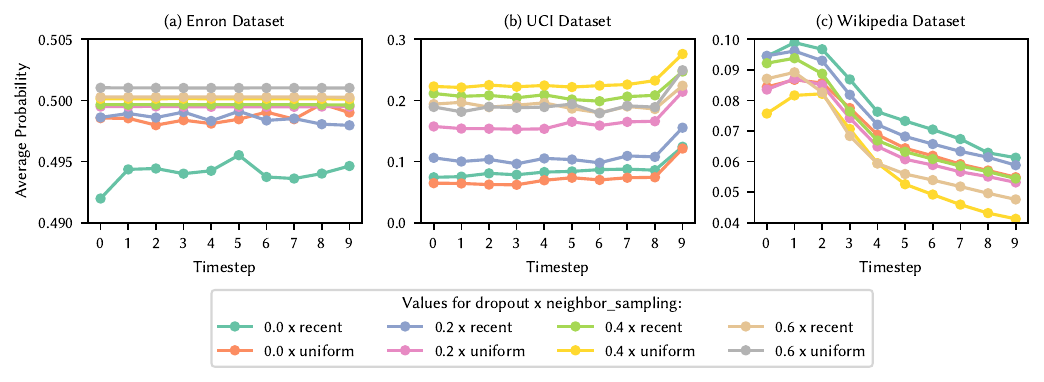}
\centering\caption{\capheader{Recency:} results for different hyperparameters for TGN. Note the very restricted y-axis range for Wikipedia.}
\label{fig:hp_recency_tgn}
\end{figure}

The recency results for \jodie{}, \tcl{}, and \tgn{} do not change qualitatively with variations of hyperparameters, as shown in Figures \ref{fig:hp_recency_jodie}, \ref{fig:hp_recency_tcl} and \ref{fig:hp_recency_tgn}.
In Tables \ref{tab:homophily_params_jodie} and \ref{tab:homophily_probs_tcl_tgn}, we also observe that different hyperparameters do not qualitatively change the homophily results for \jodie{}, \tcl{}, and \tgn{}.





\FloatBarrier

\begin{table}[h]
\caption{\capheader{Homophily}: results for different hyperparameters on the homophilic data for \jodie{}.}
\label{tab:homophily_params_jodie}
\medskip
\centering
\small
\begin{tabular}{crrrrrr}
\toprule
 & \multicolumn{3}{c}{\makecell[cc]{Avg. Edge\\Probability}}
 & \multicolumn{3}{c}{\makecell[cc]{Fraction of\\Predicted Edges}} \\
\cmidrule(lr){2-4} \cmidrule(lr){5-7}
Dropout Rate & 0-0 & 0-1 & 1-1 & 0-0 & 0-1 & 1-1 \\
\midrule
0.1 & 0.82 & 0.17 & 0.40 & 0.91 & 0.00 & 0.41 \\
0.2 & 0.82 & 0.17 & 0.42 & 0.91 & 0.00 & 0.46 \\
0.3 & 0.82 & 0.17 & 0.40 & 0.91 & 0.00 & 0.40 \\
0.4 & 0.82 & 0.17 & 0.41 & 0.88 & 0.00 & 0.42 \\
0.5 & 0.82 & 0.16 & 0.40 & 0.90 & 0.00 & 0.39 \\
0.6 & 0.80 & 0.16 & 0.39 & 0.90 & 0.00 & 0.40 \\
\bottomrule
\end{tabular}
\end{table}

\begin{table}[h]
\caption{\capheader{Homophily}: results for different hyperparameters on the homophilic data for \tcl{} and \tgn{}.}
\label{tab:homophily_probs_tcl_tgn}
\medskip
\centering
\resizebox{\linewidth}{!}{
\begin{tabular}{crrrrrrrrrrrr}
\toprule
 & \multicolumn{6}{c}{\makecell[cc]{TCL}}
 & \multicolumn{6}{c}{\makecell[cc]{TGN}} \\
\cmidrule(lr){2-7} \cmidrule(lr){8-13}
 & \multicolumn{3}{c}{\makecell[cc]{Avg. Edge\\Probability}}
 & \multicolumn{3}{c}{\makecell[cc]{Fraction of\\Predicted Edges}}
 & \multicolumn{3}{c}{\makecell[cc]{Avg. Edge\\Probability}}
 & \multicolumn{3}{c}{\makecell[cc]{Fraction of\\Predicted Edges}} \\
\cmidrule(lr){2-4} \cmidrule(lr){5-7} \cmidrule(lr){8-10} \cmidrule(lr){11-13}
Dropout $\times$ neighbor sampling & 0-0 & 0-1 & 1-1 & 0-0 & 0-1 & 1-1 & 0-0 & 0-1 & 1-1 & 0-0 & 0-1 & 1-1 \\
\midrule
0.0 $\times$ uniform 	& 0.62 & 0.40 & 0.62 & 1.00 & 0.00 & 1.00 & 0.26 & 0.14 & 0.06 & 0.31 & 0.09 & 0.04 \\
0.2 $\times$ recent 	& 0.63 & 0.40 & 0.63 & 1.00 & 0.00 & 1.00 & 0.30 & 0.19 & 0.11 & 0.33 & 0.14 & 0.08 \\
0.2 $\times$ uniform 	& 0.63 & 0.39 & 0.63 & 1.00 & 0.00 & 1.00 & 0.28 & 0.15 & 0.05 & 0.32 & 0.11 & 0.03 \\
0.4 $\times$ recent 	& 0.64 & 0.38 & 0.65 & 1.00 & 0.00 & 1.00 & 0.32 & 0.20 & 0.11 & 0.35 & 0.15 & 0.07 \\
0.4 $\times$ uniform 	& 0.64 & 0.38 & 0.65 & 1.00 & 0.00 & 1.00 & 0.27 & 0.18 & 0.09 & 0.31 & 0.12 & 0.06 \\
0.6 $\times$ recent 	& 0.66 & 0.35 & 0.67 & 1.00 & 0.00 & 1.00 & 0.32 & 0.23 & 0.14 & 0.35 & 0.16 & 0.09 \\
0.6 $\times$ uniform 	& 0.67 & 0.35 & 0.67 & 1.00 & 0.00 & 1.00 & 0.31 & 0.23 & 0.14 & 0.34 & 0.16 & 0.09 \\
\bottomrule
\end{tabular}
}
\end{table}

\FloatBarrier

\end{document}